\documentclass[10pt,journal,compsoc]{IEEEtran}
\ifCLASSOPTIONcompsoc
  \usepackage[nocompress]{cite}
\else
  \usepackage{cite}
\fi
\ifCLASSINFOpdf
\else
\fi

\usepackage{graphicx}

\usepackage{amsmath,amsfonts,bm}

\def\eqref#1{equation~\ref{#1}}

\def\1{\bm{1}}

\DeclareMathAlphabet{\mathsfit}{\encodingdefault}{\sfdefault}{m}{sl}
\SetMathAlphabet{\mathsfit}{bold}{\encodingdefault}{\sfdefault}{bx}{n}

\usepackage[export]{adjustbox}%
\usepackage{tabularx}
\usepackage{array}
\usepackage{multirow} 
\usepackage{tabularx}
\usepackage{booktabs}  
\usepackage{subcaption} 

\usepackage{color}
\usepackage[pagebackref=true,breaklinks=true,letterpaper=true,colorlinks,bookmarks=false]{hyperref}
\usepackage[dvipsnames]{xcolor}
\usepackage[misc]{ifsym}

\usepackage{ulem}
\usepackage{algorithm}
\usepackage{algpseudocode} 
\usepackage{breqn} 
\usepackage[inline]{enumitem}

\usepackage{pifont}
\newcommand{\cmark}{\ding{51}}%
\newcommand{\xmark}{\ding{55}}%

\iftrue

\newcommand{\LH}[1]{{\color{violet}{\bf LH:} #1}}

\newcommand{\SQ}[1]{{\color{black}{#1}}}
\newcommand{\YX}[1]{{\color{red}{\bf YX:} #1}}
\newcommand{\alertJW}[1]{{\color{magenta}{\bf JW:} #1}}
\else
\newcommand{\SQ}[1]{}
\newcommand{\alertJW}[1]{}
\newcommand{\YX}[1]{}
\newcommand{\CS}[1]{}
\newcommand{\LH}[1]{}
\newcommand{\FK}[1]{}
\fi

\begin{document}
%
\title{Trust your Good Friends: Source-free Domain Adaptation by Reciprocal Neighborhood Clustering}

\author{Shiqi Yang,
        Yaxing Wang$^{(\textrm{\Letter})}$, Joost van de Weijer,
        Luis Herranz, \\
        Shangling Jui, Jian Yang
\IEEEcompsocitemizethanks{\IEEEcompsocthanksitem S. Yang, J. van de Weijer and  L. Herranz are with the Computer Vision Center, Universitat Aut\`onoma de Barcelona, Barcelona 08193, Spain.\protect\\
E-mail: \{syang,joost,lherranz\}@cvc.uab.es.
\IEEEcompsocthanksitem Y. Wang, College of Computer Science, Nankai University,  China. E-mail:yaxing@nankai.edu.cn.
\IEEEcompsocthanksitem S. Jui  is  with Huawei Kirin Solution, Shanghai, China. E-mail:jui.shangling@huawei.com.
\IEEEcompsocthanksitem J. Yang  is  with College of Computer Science, Nankai University,  China. E-mail:csjyang@nankai.edu.cn.
}
\thanks{Manuscript received April 19, 2005; revised August 26, 2015.}}
%
%

\markboth{Journal of \LaTeX\ Class Files,~Vol.~14, No.~8, August~2015}%
{Shell \MakeLowercase{\textit{et al.}}: Bare Demo of IEEEtran.cls for Computer Society Journals}
%



\IEEEtitleabstractindextext{%
\begin{abstract}
Domain adaptation (DA) aims to alleviate the domain shift between source domain and target domain. Most DA methods require access to the source data, but often that is not possible (e.g. due to data privacy or intellectual property). In this paper, we address the challenging source-free domain adaptation (SFDA) problem, where the source pretrained model is adapted to the target domain in the absence of source data. Our method is based on the observation that target data, which might not align with the source domain classifier, still forms clear clusters. We capture this intrinsic structure by defining local affinity of the target data, and encourage label consistency among data with high local affinity. We observe that higher affinity should be assigned to reciprocal neighbors. 
To aggregate information with more context, we consider expanded neighborhoods with small affinity values. {Furthermore,  we consider the density around each target sample, {which can alleviate the negative impact of potential outliers.} } In the experimental results we verify that the inherent structure of the target features is an important source of information for domain adaptation. We demonstrate that this local structure can be efficiently captured by considering the local neighbors, the reciprocal neighbors, and the expanded neighborhood. Finally, we achieve state-of-the-art performance on several 2D image and 3D point cloud recognition datasets. 
\end{abstract}

}

\maketitle

\IEEEdisplaynontitleabstractindextext

%
\IEEEpeerreviewmaketitle

\IEEEraisesectionheading{\section{Introduction}\label{sec:intro}}

%
%
%
%


\IEEEPARstart{M}{ost} deep learning methods rely on training on large amounts of labeled data, while they cannot generalize well to a related yet different domain. One research direction to address this issue is Domain Adaptation (DA), which aims to transfer learned knowledge from a source to a target domain. 
Most existing DA methods demand labeled source data during the adaptation period, however, it is often not practical that source data are always accessible, such as when applied on data with privacy or property restrictions. Therefore, recently, there have emerged several works~\cite{kundu2020universal,kundu2020towards,li2020model,liang2020we} tackling a new challenging DA scenario where instead of source data only the source pretrained model is available for adapting, \textit{i.e.}, source-free domain adaptation (SFDA). Among these methods, USFDA~\cite{kundu2020universal} addresses universal DA~\cite{you2019universal} and SF~\cite{kundu2020towards} addresses open-set DA~\cite{saito2018open}. In both universal and open-set DA the label set is different for source and target domains. SHOT~\cite{liang2020we} and 3C-GAN~\cite{li2020model} are for closed-set DA where source and target domains have the same categories. 3C-GAN~\cite{li2020model} is based on target-style image generation with a conditional GAN, and SHOT~\cite{liang2020we} is based on mutual information maximization and pseudo labeling. {BAIT~\cite{yang2020unsupervised} extends MCD~\cite{saito2018maximum} to the SFDA setting.} {FR or BUFR~\cite{eastwood2021source} is based on source feature restoring.}
However, these methods ignore the intrinsic neighborhood structure of the target data in feature space which can be very valuable to tackle SFDA. {Though recent G-SFDA~\cite{yang2021generalized} consider neighborhood clustering to address SFDA, it fails to distinguish the potential noisy nearest neighbors, which may lead to performance degradation.}


\begin{figure*}[t]
	\centering
	\includegraphics[width=\textwidth]{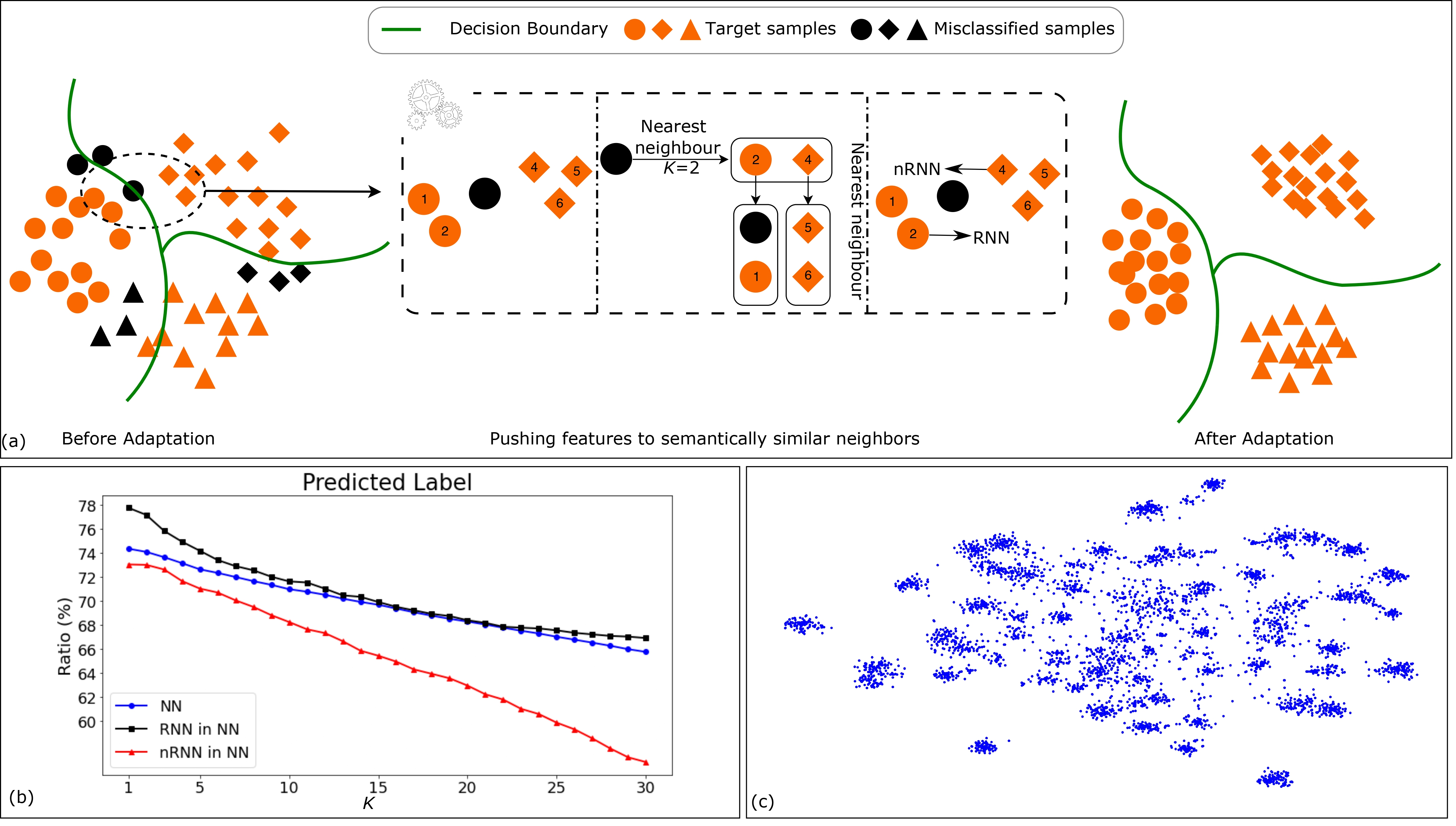}
	\caption{(\textbf{a}) Illustration of our method. In the left shows we distinguish reciprocal and non-reciprocal neighbors. The adaptation is achieved by pushed the features towards reciprocal neighbors heavily.  (\textbf{b}) Ratio of different type of nearest neighbor features of which: the \textit{predicted} label is the same as the feature, K is the number of nearest neighbors. (\textbf{c}) t-SNE visualization of target features by source model. The features in (b) and (c) are on task Ar$\rightarrow$Rw of Office-Home.    
	\vspace{-2mm}}
	\label{fig:motivation}
	\vspace{-2mm}
\end{figure*}

In this paper, we focus on source-free domain adaptation. Our main observation is that current DA methods do not exploit the intrinsic neighborhood structure of the target data. We use this term to refer to the fact that, even though the target data might have shifted in the feature space (due to the covariance shift), target data of the same class is still expected to form a cluster in the embedding space. This can be implied to some degree from the t-SNE visualization of target features on the source model which suggests that significant cluster structure is preserved (see Fig.~\ref{fig:motivation} (c)). This assumption is implicitly adopted by most DA methods, as instantiated by a recent DA work~\cite{tang2020unsupervised}.
A well-established way to assess the structure of points in high-dimensional spaces is by considering the nearest neighbors of points, which are expected to belong to the same class. However, this assumption is not true for all points; the blue curve in Fig. 1(b) shows that around 75\% of the nearest neighbors has the correct label. In this paper, we observe that this problem can be mitigated by considering reciprocal nearest neighbors (RNN); the reciprocal neighbors of a point have the point as their neighbor. Reciprocal neighbors have been studied before in different contexts~\cite{jegou2007contextual,qin2011hello,zhong2017re}. The reason why reciprocal neighbors are more trustworthy is illustrated in Fig.~\ref{fig:motivation}(a). Furthermore, Fig.~\ref{fig:motivation}(b) shows the ratio of neighbors which have the \textit{correct prediction} for different kinds of nearest neighbors. The curves show that reciprocal neighbors indeed have more chances to predict the \textit{true} label than non-reciprocal nearest neighbors (nRNN).

The above observation and analysis motivate us to assign different weights to the supervision from nearest neighbors. Our method, called Neighborhood Reciprocity Clustering (\textit{{NRC}}), achieves source-free domain adaptation by encouraging reciprocal neighbors to concord in their label prediction. 
In addition, we will also consider a weaker connection to the non-reciprocal neighbors. We define affinity values to describe the degree of connectivity between each data point and its neighbors, which is used to encourage class-consistency between neighbors. Moreover we propose to use a self-regularization to decrease the negative impact of potential noisy neighbors. 
Inspired by recent graph based methods \cite{altenburger2018monophily,chin2019decoupled,zhu2020beyond} which show that the higher order neighbors can provide relevant context, and also considering neighbors of neighbors is more likely to provide datapoints that are close on the data manifold~\cite{tenenbaum2000global}. Thus, to aggregate wider local information, we further retrieve the expanded neighbors, \textit{i.e}, neighbor of the nearest neighbors, for auxiliary supervision. 

{Though deploying the above neighborhood clustering can lead to good performance, this clustering objective may deteriorate feature representations when based on features that are outliers, since outliers typically have no semantic-similar nearest neighbors. To alleviate this circumstance, we further propose to estimate the feature density based on nearest neighbor retrieval. We then only consider those features in high density regions for clustering and give less credit to the potential outlier features. We denote this augmented version as \textbf{NRC++}.}

Our contributions can be summarized as follows, to achieve source-free domain adaptation: 
\begin{enumerate*}[label=(\Roman*)]
    \item We explicitly exploit the fact that same-class data forms cluster in the target embedding space, we do this by considering the predictions of neighbors and reciprocal neighbors. 
    \item We  show that considering an extended neighborhood of data points further improves results.
    \item  {We propose to estimate the feature density based on nearest neighbor retrieval. We then decrease the contribution of the potential outlier features in the clustering, leading to further  performance gains.}
    \item The experimental results on three 2D image datasets and one 3D point cloud dataset show that our method achieves state-of-the-art performance compared with related methods.
\end{enumerate*}

{This paper is an extension of our conference submission~\cite{yang2021exploiting}. We have extended the technical contribution, and considered new settings and a new dataset in our new version. We here summarize the main
extensions: (1) more comprehensive related works have been discussed; (2) to reduce the negative impact of outliers, we estimate the density around each data point and decrease the contribution of outliers on the clustering. {This newly proposed method, called NRC++ ,improves results on most of the experiments};
(3) we evaluate our method on additional domain adaptation settings: partial set, multi-source and multi-target domain adaptation, as well as the previous classical closed domain adaptation. (4) we conduct experiments and present results on the new challenging dataset: DomainNet~\cite{M3SDA}. }

\section{Related Work}

\textbf{Domain Adaptation.}
Most DA methods tackle domain shift by aligning the feature distributions. Early DA methods such as~\cite{long2015learning,sun2016return,tzeng2014deep} adopt moment matching to align feature distributions. And in recent years, plenty of works have emerged that achieve alignment by adversarial training. DANN~\cite{ganin2016domain} formulates domain adaptation as an adversarial two-player game. The  adversarial training of CDAN~\cite{long2018conditional} is conditioned on several sources of information. DIRT-T~\cite{shu2018dirt} performs domain adversarial training with an added term that penalizes violations of the cluster assumption. Additionally, \cite{Lee_2019_CVPR,lu2020stochastic,saito2018maximum} adopts prediction diversity between multiple learnable classifiers to achieve local or category-level feature alignment between source and target domains. AFN~\cite{Xu_2019_ICCV} shows that the erratic discrimination of target features stems from much smaller norms than those found in the source features. SRDC~\cite{tang2020unsupervised} proposes to directly uncover the intrinsic target discrimination via discriminative clustering to achieve adaptation. {More related, \cite{pan2020exploring} resorts to K-means clustering for open-set adaptation while considering global structure. Our method instead only focuses on nearest neighbors (local structure) for source-free adaptation}. The most relevant paper to ours is DANCE~\cite{saito2020universal}, which is for universal domain adaptation and based on neighborhoods clustering. But they compute the entropy of instance discrimination~\cite{wu2018unsupervised} between all features, thus the non-local neighborhood clustering. In our method, we encourage prediction consistency between only a few semantically close neighbors. {There are also several different domain adaptation paradigms, such as partial-set domain adaptation~\cite{zhang2018importance,li2020deep_pda,cao2019learning,liang2020balanced} where the label space of the source domain contains the one of the target domain, open-set domain adaptation~\cite{saito2018open,liu2019separate} where the label space of the source domain is included in the one of the target domain, universal domain adaptation~\cite{you2019universal,saito2020universal} where there exist both domain specific and domain shared categories, multi-source domain adaptation~\cite{SImpAl,CMSDA,DRT,STEM} where there are multiple different labeled source domains for training, and multi-target domain adaptation~\cite{CGCT,nguyen2020unsupervised} where there are multiple unlabeled target domains for training and evaluation.}

\noindent \textbf{Source-free Domain Adaptation.}
Source-present methods need supervision from the source domain during adaptation. Recently, there are several methods investigating source-free domain adaptation. For the closed-set DA setting, BAIT~\cite{yang2020unsupervised} extends MCD~\cite{saito2018maximum} to source-free setting, and SHOT~\cite{liang2020we} proposes to fix the source classifier and match the target features to the fixed classifier by maximizing mutual information and a proposed pseudo label strategy {which considers global structure}. SHOT++~\cite{liang2021source} uses both the self-supervised and the semi-supervised learning techniques for further improving SHOT. \SQ{And several other methods address SFDA by generating features}, 3C-GAN~\cite{li2020model} synthesizes labeled target-style training images based on the conditional GAN to provide supervision for adaptation, while SFDA~\cite{liu2021source} tackles the segmentation task by synthesizing fake source samples. \SQ{Along with attention mechanism to avoid forgetting on the source domain}, G-SFDA~\cite{yang2021generalized} propose neighborhood clustering which enforces prediction consistency between local neighbors. \SQ{Based on Instance Discrimination~\cite{wu2018unsupervised}, HCL~\cite{huang2021model} adopts features from current and historical models to cluster features, as well as a generated pseudo label conditioned on historical consistency.} Recently, FR or BUFR~\cite{eastwood2021source} proposes to restore the source features to address SFDA, by adapting the feature-extractor with only target data such that the approximate feature distribution under the target data realigns with that saved distribution on the source. USFDA~\cite{kundu2020universal} and FS~\cite{kundu2020towards} explore source-free universal DA~\cite{you2019universal} and open-set DA~\cite{saito2018open}, and they propose to synthesize extra training samples to make the decision boundary compact, thereby allowing to recognize the open classes. DECISION~\cite{DECISION} addresses source-free multi-source domain adaptation where the model is first pretrained on multiple labeled source domains and then adapted to the target domain without access to source data anymore. \SQ{Recently \cite{yang2022attracting} proposes a simple clustering objective to achieve adaptation by clustering features. To address the imbalance issue in the feature clustering stage, \cite{qu2022bmd} proposes a dynamic pseudo labeling strategy.} And recently there also emerge several works on test-time adaptation~\cite{wang2020fully,wang2022continual,choi2022improving,niu2022efficient,chen2022contrastive} which can be actually regarded as an online source-free domain adaptation task, while the training and evaluation protocol are different. We will not detail them in this paper.



\noindent \textbf{Graph Clustering.} Our method shares some similarities with graph clustering work such as~\cite{sarfraz2019efficient,wang2019linkage,yang2020learning,yang2019learning} by utilizing neighborhood information. However, our methods are fundamentally different. Unlike those works which require labeled data to train the graph network for estimating the affinity, we instead adopt reciprocity to assign affinity. 

\section{Method}

\noindent \textbf{Notation.}
We denote the labeled source domain data with $n_s$ samples as $\mathcal{D}_s = \{(x_i^s,y^s_i)\}_{i=1}^{n_s}$, where the $y^s_i$ is the corresponding label of $x_i^s$, and the unlabeled target domain data with $n_t$ samples as $\mathcal{D}_t=\{x_j^t\}_{j=1}^{n_t}$. Both domains have the same $C$ classes (closed-set setting). Under the SFDA setting $\mathcal{D}_s$ is only available for model pretraining. Our method is based on a neural network, which we split into two parts: a feature extractor $f$, and a classifier $g$. The feature output by the feature extractor is denoted as $\bm{z}(x)=f\left(x\right)$, the output of network is denoted as $p(x)=\delta(g(z)) \in \mathcal{R}^C$ where $\delta$ is the softmax function, for readability we will omit the input and use $\bm{z}, p$ in the following sections. 

\noindent \textbf{Overview.} We assume that the source pretrained model has already been trained. As discusses in the introduction, the target features output by the source model form clusters. We exploit this intrinsic structure of the target data for SFDA by considering the neighborhood information, and the adaptation is achieved with the following objective:
\begin{eqnarray}\label{eq:raw}
    \mathcal{L}=-\frac{1}{n_t}\sum_{x_i\in \mathcal{D}_t}\sum_{x_j \in \text{Neigh}(x_i)}  \frac{D_{sim}(p_i,p_j)}{D_{dis}(x_i,x_j)}
\end{eqnarray}
where the $\text{Neigh}(x_i)$ means the nearest neighbors of $x_i$, $D_{sim}$ computes the similarity between predictions, and $D_{dis}$ is a constant measuring the semantic distance (dissimilarity) between data. The principle behind the objective is to push the data towards their semantically close neighbors by encouraging similar predictions. In the next sections, we will define $D_{sim}$ and $D_{dis}$.

\subsection{Encouraging Class-Consistency with Neighborhood Affinity} 
To achieve adaptation without source data, we use the prediction of the nearest neighbor to encourage prediction consistency. 
{While the target features computed with the source model are not necessarily discriminative, meaning some neighbors belong to different classes and will provide incorrect supervision. To decrease the potentially negative impact of those neighbors, we propose to weigh the supervision from neighbors according to the connectivity (semantic similarity)}.
We define \textit{affinity} values to signify the connectivity between the neighbor and the feature, which corresponds to the $\frac{1}{D_{dis}}$ in Eq.~\ref{eq:raw} indicating the semantic similarity. 

To retrieve the nearest neighbors for batch training, similar to \cite{saito2020universal,wu2018unsupervised,zhuang2019local}, we build two memory banks: $\mathcal{F}$ 
stores all target features, and $\mathcal{S}$ stores corresponding prediction scores: 
\begin{eqnarray}\label{def:banks}
    \mathcal{F}=[\bm{z}_1,\bm{z}_2,\dots,\bm{z}_{n_t}] \text{ and }
    \mathcal{S}=[p_1,p_2,\dots,p_{n_t}]
\end{eqnarray}
We use the cosine similarity for nearest neighbors retrieving. 
The difference between ours and \cite{saito2020universal,wu2018unsupervised} lies in the fact that we utilize the memory bank to retrieve nearest neighbors while \cite{saito2020universal,wu2018unsupervised} adopts the memory bank to compute the instance discrimination loss. Before every mini-batch training, we simply update the old items in the memory banks corresponding to current mini-batch. Note that updating the memory bank is only done to replace the old low-dimension vectors with new ones computed by the model, and does not require any additional computation.


\begin{figure}[t]
	\centering
	\includegraphics[width=.49\textwidth]{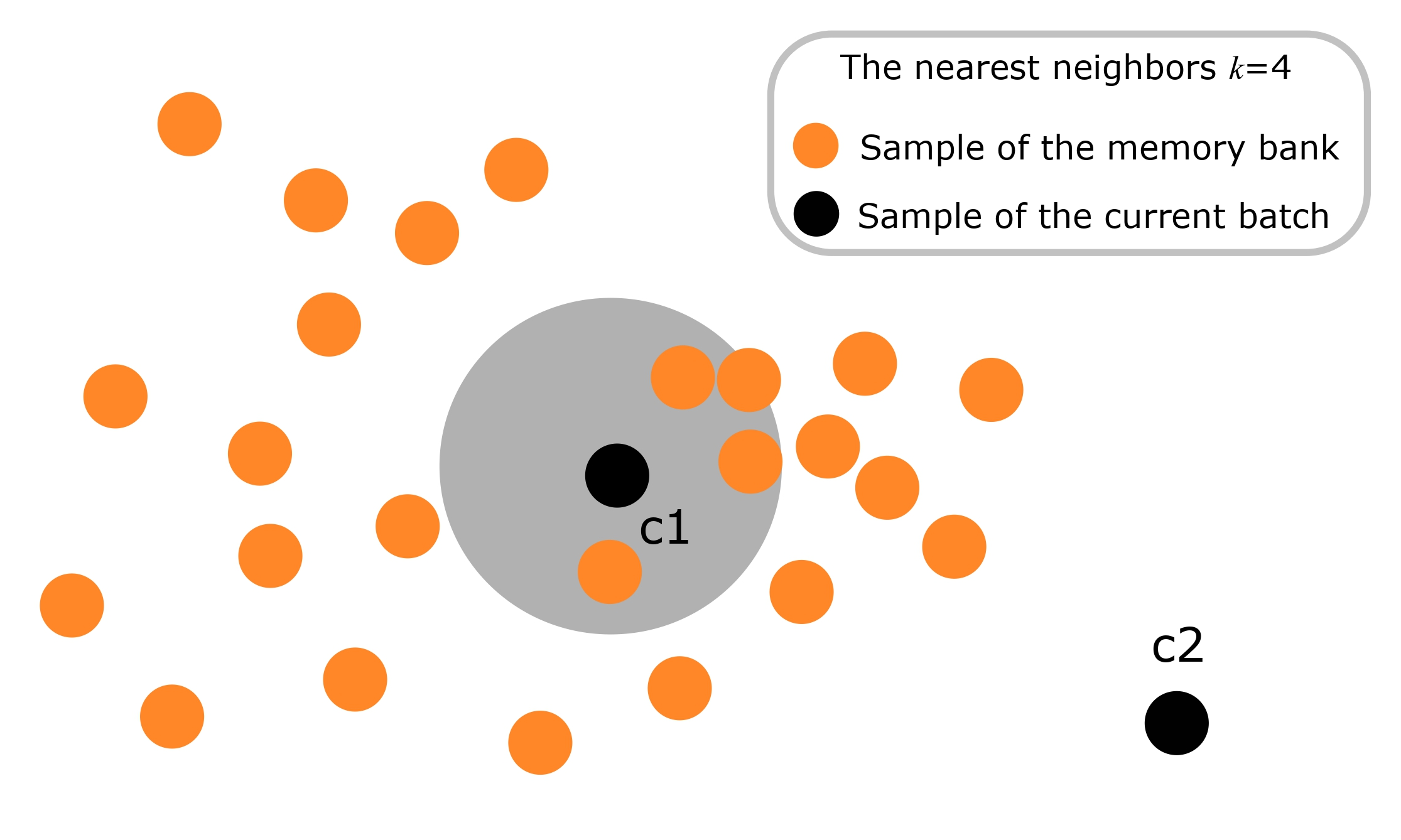}
	\caption{{Illustration of Neighborhood Density for Outlier Detection. C1 is not an outlier as a few nearest neighbors of several features in the memory bank contain C1, while C2 is an outlier and should not be included during training, since no features in the memory banks contain it as nearest neighbor.} 
	}
	\label{fig:outlier}
\end{figure}

We then use the prediction of the neighbors to supervise the training weighted by the affinity values, with the following objective adapted from Eq.~\ref{eq:raw}:
\begin{eqnarray}\label{eq:nn}
    \mathcal{L}_{\mathcal{N}}=-\frac{1}{n_t}\sum_{i}\sum_{k\in \mathcal{N}_K^i} A_{ik} \mathcal{S}_k^\top p_i
\end{eqnarray}
where we use the dot product to compute the similarity between predictions, corresponding to $D_{sim}$ in Eq.\ref{eq:raw}, the $k$ is the index of the $k$-th nearest neighbors of $\bm{z}_i$, $\mathcal{S}_k$ is the $k$-th item in memory bank $\mathcal{S}$, $A_{ik}$ is the affinity value of $k$-th nearest neighbors of feature $\bm{z}_i$. Here the $\mathcal{N}_K^i$ is the index set\footnote{All indexes are in the same order for the dataset and memory banks.} of the $K$-nearest neighbors of feature $\bm{z}_i$. Note that all neighbors are retrieved from the feature bank $\mathcal{F}$. With the affinity value as weight, this objective pushes the features to their neighbors with strong connectivity and to a lesser degree to those with weak connectivity. 

To assign larger affinity values to semantic similar neighbors, we divide the nearest neighbors retrieved into two groups: reciprocal nearest neighbors (RNN) and non-reciprocal nearest neighbors (nRNN). The feature $\bm{z}_j$ is regarded as the RNN of the feature $\bm{z}_i$ if it meets the following condition:
\begin{eqnarray}\label{def:rnn}
    j \in \mathcal{N}_K^i \wedge i \in \mathcal{N}_M^j
\end{eqnarray}
Other neighbors which do not meet the above condition are nRNN. Note that the normal definition of reciprocal nearest neighbors~\cite{qin2011hello} applies $K=M$, while in this paper $K$ and $M$ can be different.
We find that reciprocal neighbors have a higher potential to belong to the same cluster as the feature (Fig.~\ref{fig:motivation}(b)). Thus, we assign a high affinity value to the RNN features. Specifically for feature $\bm{z}_i$, the affinity value of its $j$-th K-nearest neighbor is defined as:
\begin{eqnarray}\label{def:a}
    A_{i, j} =
\begin{cases}
1 & \text{if }j \in \mathcal{N}_K^i \wedge i \in \mathcal{N}_M^j    \\ 
r & \text{otherwise},
\end{cases} 
\end{eqnarray}
%
where $r$ is a hyperparameter. If not specified $r$ is set to 0.1.

To further reduce the potential impact of noisy neighbors in $\mathcal{N}_K$, which belong to the different class but still are RNN, we propose a simply yet effective way dubbed \textit{self-regularization}, that is, to not ignore the current prediction of ego feature:
\begin{eqnarray}\label{eq:self}
    \mathcal{L}_{self}=-\frac{1}{n_t}\sum_{i}^{n_t} \mathcal{S}_i^\top p_i
\end{eqnarray}
where $\mathcal{S}_i$ means the stored prediction in the memory bank, note this term is a \textit{constant vector} and is identical to the $p_i$ since we update the memory banks before the training, \SQ{here the loss is only back-propagated for variable $p_i$.}

To avoid the degenerated solution~\cite{ghasedi2017deep,shi2012information} where the model predicts all data as some specific classes (and does not predict other classes for any of the target data), we encourage the prediction to be balanced. We adopt the prediction diversity loss which is widely used in clustering~\cite{ghasedi2017deep,gomes2010discriminative,jabi2019deep} and also in several domain adaptation works~\cite{liang2020we,shi2012information,tang2020unsupervised}:
\begin{eqnarray}\label{eq:div}
\mathcal{L}_{div}=\sum_{c=1}^{C} \textrm{KL}(\bar{p}_c||q_c), &\text{with}& \bar{p}_c=\frac{1}{n_t}\sum_{i} p_{i}^{(c)}  ,\\
&\textrm{and}& q_{\{c=1,..,C\}}= \frac{1}{C} \notag
\end{eqnarray}
where the $p_{i}^{(c)}$ is the score of the $c$-th class and $\bar{p}_c$ is the empirical label distribution, it represents the predicted possibility of class $c$ and q is a uniform distribution.

\subsection{Expanded Neighborhood Affinity}\label{sec:expand}

\begin{figure}[t]
	\centering
	\includegraphics[width=\columnwidth]{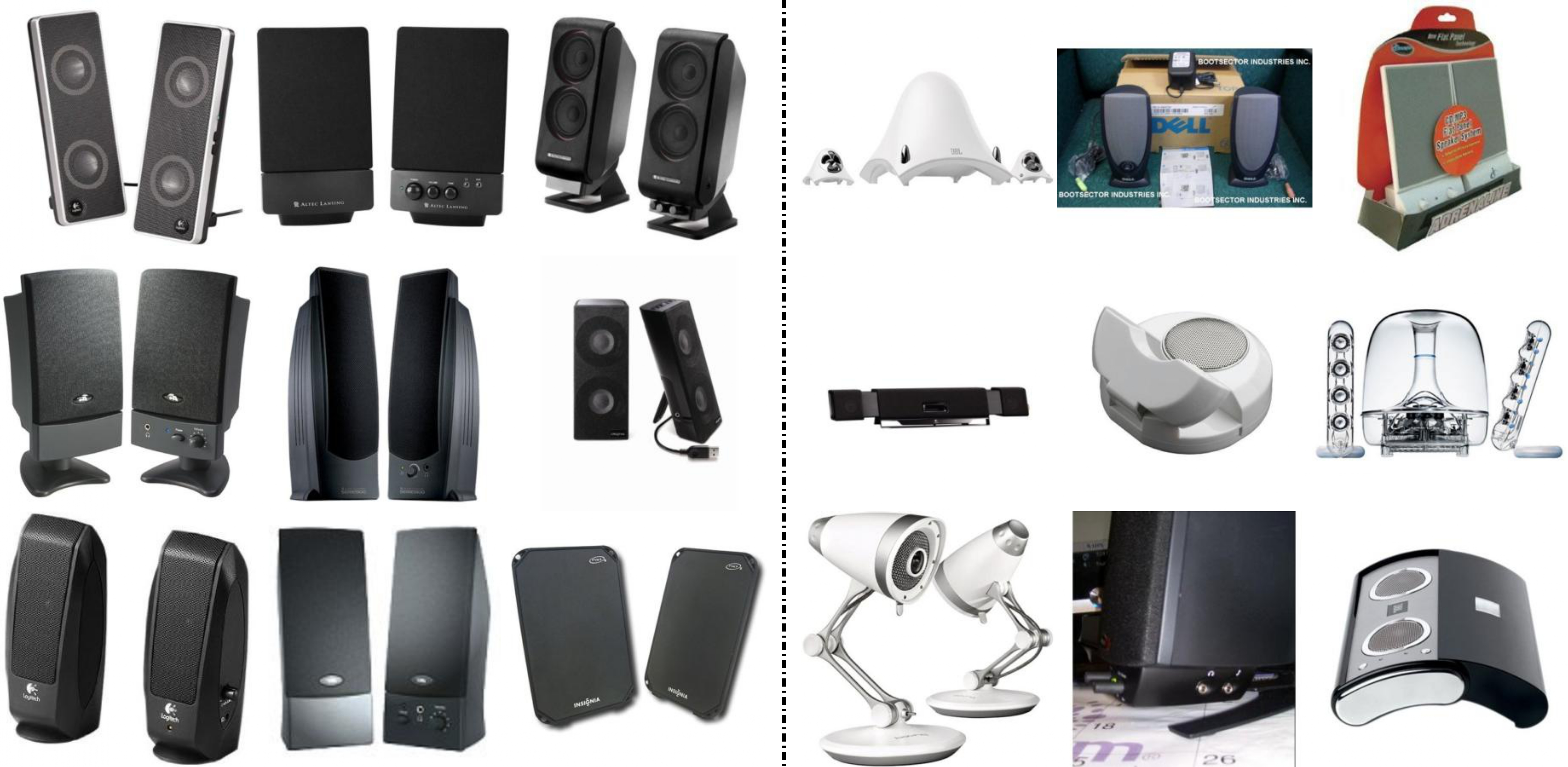}
	\caption{{Examples located in high density (left) and lower density (right). The examples are from VisDA-C~\cite{peng2017visda}.}}
	\label{fig:density_examples}
\end{figure}


As mentioned in Sec.~\ref{sec:intro}, a simple way to achieve the aggregation of more information is by considering more nearest neighbors. However, a drawback is that larger neighborhoods are expected to contain more datapoint from multiple classes, defying the purpose of class consistency. A better way to include more target features is by considering the $M$-nearest neighbor of each neighbor in $\mathcal{N}_K$ of $\bm{z}_i$ in Eq.~\ref{def:rnn}, \textit{i.e.}, the expanded neighbors. These target features are expected to be closer on the target data manifold than the features that are included by considering a larger number of nearest neighbors~\cite{tenenbaum2000global}.
The expanded neighbors of feature $\bm{z}_i$ are defined as $E_M(\bm{z}_i) = \mathcal{N}_M(\bm{z}_j)\ \forall j \in \mathcal{N}_K(\bm{z}_i)$, \textit{note that $E_M(\bm{z}_i)$ is still an index set and $i \text{\textit{ (ego feature)}} \notin E_M(\bm{z}_i)$}. 
We directly assign a small affinity value $r$ to those expanded neighbors, since they are further than nearest neighbors and may contain noise.
We utilize the prediction of those expanded neighborhoods for training:
\begin{eqnarray}\label{eq:nn2}
    \mathcal{L}_{E}=-\frac{1}{n_t}\sum_{i}\sum_{k\in \mathcal{N}_K^i} \sum_{m\in {E}_M^k}r  \mathcal{S}_m^\top p_i
\end{eqnarray}
where ${E}_M^k$ contain the $M$-nearest neighbors of neighbor $k$ in $\mathcal{N}_K$. 

Although the affinity values of all expanded neighbors are the same, it does not necessarily mean that they have equal importance. Taking a closer look at the expanded neighbors $E_M(\bm{z}_i)$, some neighbors will show up more than once, for example $\bm{z}_m$ can be the nearest neighbor of both $\bm{z}_h$ and $\bm{z}_j$ where $h,j \in \mathcal{N}_K(\bm{z}_i)$, and the nearest neighbors can also serve as expanded neighbor. It implies that those neighbors form compact cluster, and we posit that those duplicated expanded neighbors have potential to be semantically closer to the ego-feature $\bm{z}_i$. Thus, we do not remove duplicated features in $E_M(\bm{z}_i)$, as those can lead to actually larger affinity value for those expanded neighbors. This is one advantage of utilizing expanded neighbors instead of more nearest neighbors, we will verify the importance of maintaining  the duplicated features in the experimental section.


{
\begin{algorithm}[t]
 	\footnotesize
	\caption{Neighborhood Reciprocity Clustering for Source-free Domain Adaptation}
	\label{alg:snr}
	\begin{algorithmic}[1]
		\Require $\mathcal{D}_s$ (only for source model training), $\mathcal{D}_t$ 
		\State Pre-train model on $\mathcal{D}_s$
		\State Build feature bank $\mathcal{F}$ and score bank $\mathcal{S}$ for $\mathcal{D}_t$
		\While{Adaptation}
		\State Sample batch $\mathcal{T}$ from $\mathcal{D}_t$ 
		\State Update $\mathcal{F}$ and $\mathcal{S}$ corresponding to current batch $\mathcal{T}$
		\State Retrieve nearest neighbors $\mathcal{N}$ for each of $\mathcal{T}$
		\State Compute affinity values $A$ {and $B$} \Comment{Eqs.\ref{def:a}, \ref{def:density_a}}  
		\State Retrieve expanded neighborhoods $E$ for each of $\mathcal{N}$ 
		\State Compute loss and update the model\Comment{Eq.~\ref{eq:final}}
		\EndWhile 
	\end{algorithmic}
\end{algorithm}}

\subsection{{Neighborhood Density for Outlier Detection}}\label{sec:dense}
{In previous sections, we directly deploy nearest neighborhood clustering for source-free domain adaptation. However, it may deteriorate the feature representation when the features in the current batch exist as outliers. An outlier typically will not be retrieved as nearest neighbor of other features, and more importantly, whether the retrieved nearest neighbors of the outlier belong to the same semantic cluster is often unsure. Thus, in this section, we propose to filter out the potential outlier features, and exclude them in the objective computation.} 


{To find those outlier features, we resort to nearest neighbor retrieval of the features in the memory bank. For each feature $z_j$ in the memory bank, we retrieve its $U$ nearest neighbors. The density of the feature $i$ can be estimated by counting \textit{how many} samples have $i$ as its nearest neighbor. This is given by $||\mathcal{D}(i)||$ where }
\begin{align}
\mathcal{D}(i):=&{\{j|i\in \mathcal{N}_U^j\}}
\label{def:density}.
\end{align}
The more samples in $\mathcal{D}(i)$, the larger the density around the sample $x_i$.

{Having identified the outliers, we can now proceed and exclude from the clustering. We therefore define $B$, similar to Eq.~\ref{def:a}, to be:} 
\begin{eqnarray}\label{def:density_a}
    B_{i, j} =
\begin{cases}
1 & \text{if }j \in (\mathcal{D}(i) \bigcap  \mathcal{N}_V^i)    \\ 
r & \text{otherwise},
\end{cases} 
\end{eqnarray}
and the loss is given by: 
\begin{eqnarray}\label{eq:density_nn}
    \mathcal{L}_{\mathcal{D}}=-\frac{1}{n_t}\sum_{i}\sum_{j\in \mathcal{D}(i)} B_{ij} \mathcal{S}_j^\top p_i.
\end{eqnarray}
%
%
%
{Here the $r$ is a hyperparameter, if not specified it is set to 0.1. We identify the method that includes this loss with \textit{NRC++}.

As illustrated in Fig.~\ref{fig:outlier}, in Eq.~\ref{def:density_a} when the feature $i$ is an outlier, which means $\mathcal{D}_i$ is the empty set, it will be excluded in Eq.~\ref{eq:density_nn}. }
{If the feature $x_i$ is not an outlier, then Eq.~\ref{eq:density_nn} will have a similar meaning as Eq.~\ref{eq:nn}. Note that in Eq.~\ref{def:density_a} the second summation is over $D(i)$ which is different from Eq.~\ref{eq:nn}. As a result, both losses are considering different neighbors. When applied jointly they constitute a clustering algorithm that is less sensitive to outliers. And Fig~\ref{fig:density_examples} shows the retrieved samples which are located in higher density (larger $||\mathcal{D}(i)||$) and lower density regions (smaller $||\mathcal{D}(i)||$).}



\begin{table*}[t]
\centering
\caption{Accuracies (\%) on Office-31 for ResNet50-based methods.}
\begin{center}
		\scalebox{1}{
	\setlength{\tabcolsep}{5mm}{
	\resizebox{\textwidth}{!}{	\begin{tabular}{l|c|ccccccc}
					\hline
					Method & \multicolumn{1}{|c|}{SF}&A$\rightarrow$D & A$\rightarrow$W & D$\rightarrow$W & W$\rightarrow$D & D$\rightarrow$A & W$\rightarrow$A & Avg \\
					\toprule
					MCD \cite{saito2018maximum}& \xmark&{92.2} & {88.6} & {98.5} & \textbf{100.0} & {69.5} & {69.7} & {86.5} \\	
					CDAN \cite{long2018conditional}& \xmark & {92.9} & {94.1} & {98.6} & \textbf{100.0} & {71.0} & {69.3} & {87.7} \\	
					CBST~\cite{zou2018unsupervised}  &\xmark&86.5  &87.8  &98.5 &\textbf{100.0}  &70.9  &71.2  &85.8 \\
					MDD~\cite{zhang2019bridging}& \xmark& 90.4 &90.4  &98.7 &99.9 &75.0  &73.7 &88.0\\
					MDD+IA~\cite{jiang2020implicit}&\xmark&92.1  &90.3  &98.7 &99.8 &{75.3}  &74.9 &{88.8}\\
					{BNM}~\cite{cui2020towards}& \xmark   &90.3  &91.5  &98.5 &\textbf{100.0} &70.9  &71.6 &87.1\\	
				
					DMRL~\cite{wu2020dual}& \xmark&93.4  &90.8  &{99.0} &\textbf{100.0} &{73.0}  &71.2 &87.9\\
					BDG~\cite{yang2020bi}& \xmark&93.6  &93.6  &{99.0} &\textbf{100.0} &{73.2}  &72.0 &88.5\\
					MCC~\cite{jin2019minimum}& \xmark&{95.6}  &	{95.4}  &98.6 &100.0 &{72.6}  &73.9 &{89.4}\\
					SRDC~\cite{tang2020unsupervised}& \xmark&{95.8}  &{95.7}  &{99.2} &100.0 &{76.7}  &77.1 &{90.8}\\
					RWOT~\cite{xu2020reliable}& \xmark&{94.5}  &{95.1}  &\textbf{99.5} &100.0 &\textbf{77.5}  &77.9 &{90.8}\\
					RSDA~\cite{gu2020spherical}& \xmark&{95.8}  &\textbf{96.1}  &{99.3} &  \textbf{100.0}& 77.4 & \textbf{78.9} &\textbf{91.1}\\
					\toprule
					
					SHOT~\cite{liang2020we}& \cmark & {94.0} &  90.1 &  98.4 & 99.9  & 74.7 & 74.3 & {88.6}\\
					3C-GAN~\cite{li2020model}& \cmark & {92.7} &  93.7 &  98.5 & 99.8  & {75.3} & {77.8} & {89.6}\\
					{HCL}~\cite{huang2021model}& \cmark &{94.7}	&{92.5} & {98.2}  &100.0	&{75.9}	&{77.7}	&{89.8}\\
					
					{\textbf{NRC}}& \cmark &\textbf{96.0}	&{90.8} & {99.0}  &\textbf{100.0}	&{75.3}	&{75.0}	&{89.4}\\
					{\textbf{NRC++}}& \cmark &{95.9}	&{91.2} & {99.1}  &\textbf{100.0}	&{75.5}	&{75.0}	&{89.5}\\
					\hline
			\end{tabular}
			}
		}}
		\end{center}
				\label{tab:office31}
\end{table*}

\noindent \textbf{Final objective.}
Our method, called {Neighborhood Reciprocity Clustering} {({NRC} and {NRC++})},  is illustrated in Algorithm.~\ref{alg:snr}. The final objective for adaptation is:
\begin{eqnarray}\label{eq:final}
    \mathcal{L}=\mathcal{L}_{\mathcal{N}}+\mathcal{L}_{\mathcal{D}}+\mathcal{L}_E+\mathcal{L}_{self} +  \lambda_{div}\mathcal{L}_{div},
\end{eqnarray}
{where hyper-parameter $\lambda_{div}$ aims to balance $\mathcal{L}_{div}$. In our experiment, we gradually reduce  $\lambda_{div}$ value with weight decay, since we consider that $\mathcal{L}_{div}$ plays a more important role at the early training stage since the target data are probably disorderly clustered together. We reduce the influence of $\mathcal{L}_{div}$ when the target data starts to form semantic clusters. }



\begin{table*}[t]
\caption{Accuracies (\%) on Office-Home for ResNet50-based methods.}
	\begin{center}
		\linespread{1.0}
	\setlength{\tabcolsep}{2mm}{
			\resizebox{\textwidth}{!}{%
				\begin{tabular}{l|c|ccccccccccccc}
					\hline
					Method & \multicolumn{1}{|c|}{SF}& Ar$\rightarrow$Cl & Ar$\rightarrow$Pr & Ar$\rightarrow$Rw & Cl$\rightarrow$Ar & Cl$\rightarrow$Pr & Cl$\rightarrow$Rw & Pr$\rightarrow$Ar & Pr$\rightarrow$Cl & Pr$\rightarrow$Rw & Rw$\rightarrow$Ar & Rw$\rightarrow$Cl & Rw$\rightarrow$Pr & \textbf{Avg} \\
					\toprule
					ResNet-50 \cite{he2016deep} &\xmark& 34.9 & 50.0 & 58.0 & 37.4 & 41.9 & 46.2 & 38.5 & 31.2 & 60.4 & 53.9 & 41.2 & 59.9 & 46.1 \\
					DAN \cite{long2015learning} &\xmark& 43.6 & 57.0 & 67.9 & 45.8 & 56.5 & 60.4 & 44.0 & 43.6 & 67.7 & 63.1 & 51.5 & 74.3 & 56.3 \\
					DANN \cite{ganin2016domain} &\xmark& 45.6 & 59.3 & 70.1 & 47.0 & 58.5 & 60.9 & 46.1 & 43.7 & 68.5 & 63.2 & 51.8 & 76.8 & 57.6 \\
					MCD \cite{saito2018maximum}&\xmark &48.9	&68.3	&74.6	&61.3	&67.6	&68.8	&57.0	&47.1	&75.1	&69.1	&52.2	&79.6	&64.1 \\
					CDAN \cite{long2018conditional}&\xmark &50.7	&70.6	&76.0	&57.6	&70.0	&70.0	&57.4	&50.9	&77.3	&70.9	&56.7	&81.6	&65.8 \\
					SAFN~\cite{Xu_2019_ICCV}&\xmark &52.0&	71.7&76.3&64.2&69.9&71.9&63.7&51.4&	77.1&70.9&57.1&81.5&67.3\\
					Symnets \cite{zhang2019domain} &\xmark& 47.7 & 72.9 & 78.5 & {64.2} & 71.3 & 74.2 & {64.2}  & 48.8 & {79.5} & {74.5} & 52.6 & {82.7} & 67.6 \\
					MDD \cite{zhang2019bridging}&\xmark & 54.9 & 73.7 & 77.8 & 60.0 & 71.4 & 71.8 & 61.2 & {53.6} & 78.1 & 72.5 & \textbf{60.2} & 82.3 & 68.1 \\
					TADA \cite{wang2019transferable1} &\xmark& 53.1 & 72.3 & 77.2 & 59.1 & 71.2 & 72.1 & 59.7 & {53.1} & 78.4 & 72.4 & {60.0} & 82.9 & 67.6 \\
					
					MDD+IA~\cite{jiang2020implicit}&\xmark& 56.0 & 77.9 & 79.2 & 64.4 & 73.1 & 74.4 & 64.2 & {54.2} & 79.9 & 71.2 & {58.1} & 83.1 & 69.5 \\
					{BNM} \cite{cui2020towards}&\xmark & 52.3 & 73.9 & {80.0} & 63.3 & {72.9} & {74.9} & 61.7 & {49.5} & {79.7} & 70.5 & {53.6} & 82.2 & 67.9 \\
					AADA+CCN \cite{yangmind} &\xmark& 54.0 & 71.3 & {77.5} & 60.8 & {70.8} & {71.2} & 59.1 & {51.8} & {76.9} & 71.0 & {57.4} & 81.8 & 67.0 \\
					BDG \cite{yang2020bi}&\xmark & 51.5 & 73.4 & {78.7} & 65.3 & {71.5} & {73.7} & 65.1 & {49.7} & {81.1} & 74.6 & {55.1} & 84.8 & 68.7 \\
					SRDC \cite{tang2020unsupervised}&\xmark & 52.3 & 76.3 & {81.0} & \textbf{69.5} & {76.2} & {78.0} & \textbf{68.7} & {53.8} & {81.7} & \textbf{76.3} & {57.1} & {85.0} & {71.3} \\
					RSDA~\cite{gu2020spherical}&\xmark&53.2&77.7&81.3&66.4&74.0&76.5&67.9&53.0&82.0&75.8&57.8&85.4&70.9\\
					\toprule
					SHOT~\cite{liang2020we}&\cmark &{57.1} & {78.1} & 81.5 & {68.0} & {78.2} & {78.1} & {67.4} & 54.9 & {82.2} & 73.3 & 58.8 & {84.3} & {71.8}  \\
				
				{\textbf{NRC}}&\cmark &{57.7} 	&{80.3} 	&\textbf{82.0} 	&{68.1} 	&{79.8} 	&{78.6} 	&{65.3} 	&{56.4} 	&{83.0} 	&71.0	&{58.6} 	&{85.6} 	&{72.2} \\
		    {\textbf{NRC++}}&\cmark &\textbf{57.8} 	&\textbf{80.4} 	&{81.6} 	&{69.0} 	&\textbf{80.3} 	&\textbf{79.5} 	&{65.6} 	&\textbf{57.0} 	&\textbf{83.2} 	&72.3	&{59.6} 	&\textbf{85.7} 	&\textbf{72.5} \\
					\hline
		\end{tabular}}
		}
				\label{tab:home}
	\end{center}
\end{table*}

\section{Experiments}\label{sec:exp}
\noindent \textbf{Datasets.} We use three 2D image benchmark datasets and a 3D point cloud recognition dataset. \textbf{Office-31}~\cite{saenko2010adapting} contains 3 domains (Amazon(\textbf{A}), Webcam(\textbf{W}), DSLR(\textbf{D})) with 31 classes and 4,652 images. \textbf{Office-Home}~\cite{venkateswara2017deep} contains 4 domains (Real(\textbf{Rw}), Clipart(\textbf{Cl}), Art(\textbf{Ar}), Product(\textbf{Pr})) with 65 classes and a total of 15,500 images. \textbf{VisDA}~\cite{peng2017visda} is a more challenging dataset, with 12-class synthetic-to-real object recognition tasks, its source domain contains of 152k synthetic images while the target domain has 55k real object images. {\textbf{DomainNet}~\cite{M3SDA}
is the most challenging with six distinct domains (345 classes and about 0.6 million images ): Clipart (\textbf{C}), Real (\textbf{R}), Infograph (\textbf{I}), Painting (\textbf{P}), Sketch (\textbf{S}), and Quickdraw (\textbf{Q})}. \textbf{PointDA-10}~\cite{qin2019pointdan} is the first 3D point cloud benchmark specifically designed for domain adaptation, it has 3 domains with 10 classes, denoted as ModelNet-10, ShapeNet-10 and ScanNet-10, containing approximately 27.7k training and 5.1k testing images together.

\noindent \textbf{Evaluation.} We compare with existing source-present and source-free DA methods. 
\textit{All results are the average of three random runs.} \textbf{SF} in the tables denotes source-free. {In this paper, we do not compare with shot++~\cite{liang2021source}, which mainly uses extra self-supervised and semi-supervised learning procedures to improve the generalizability of the model, thus further improving the final performance. }

\noindent \textbf{Model details.} For fair comparison with related methods, we also adopt the backbone of ResNet-50~\cite{he2016deep} for Office-Home and  ResNet-101 for VisDA, and PointNet~\cite{qi2017pointnet} for PointDA-10. Specifically, for 2D image datasets, we use the same network architecture as SHOT~\cite{liang2020we}, \textit{i.e.}, the final part of the network is: $\text{fully connected layer}\ -\ \text{Batch Normalization~\cite{ioffe2015batch}}\ -\ \text{fully connected layer}\ \text{with weight normalization~\cite{salimans2016weight}}$.
And for PointDA-10~\cite{qi2017pointnet}, we use the code released by the authors for fair comparison with PointDAN~\cite{qi2017pointnet}, and only use the backbone without any of their proposed modules. To train the source model, we also adopt label smoothing as SHOT does. We adopt SGD with momentum 0.9 and batch size of 64 for all 2D datasets, and Adam for PointDA-10. The learning rate for Office-31 and Office-Home is set to 1e-3 for all layers, except for the last two newly added fc layers, where we apply 1e-2. Learning rates are set 10 times smaller for VisDA. Learning rate for PointDA-10 is set to 1e-6. We train 30 epochs for Office-31 and Office-Home while 15 epochs for VisDA, and 100 for PointDA-10. For the number of nearest neighbors (K, U, V) and expanded neighborhoods (M), we use {3, 20, 5, 2} for Office-31, Office-Home and PointDA-10, since VisDA is much larger we set K, M to 5, {and U, V to 20, 5}. \SQ{As for the decay factor $\mathcal{L}_{div}$ in Eq.~\ref{eq:final}, it is defined as $(1 + 10 \times \frac{current\_iter}{max\_iter})^{-1}$.}

\begin{table*}[t]
\caption{Accuracies (\%) on VisDA-C (Synthesis $\to$ Real) for ResNet101-based methods.}
	\begin{center}
		\linespread{1.0}
			\resizebox{\textwidth}{!}{%
			\begin{tabular}{l|c|ccccccccccccc}
			\hline
			Method&\multicolumn{1}{|c|}{SF} & plane & bcycl & bus & car & horse & knife & mcycl & person & plant & sktbrd & train & truck & Per-class \\
			\toprule
			ResNet-101 \cite{he2016deep}&\xmark  & 55.1 & 53.3 & 61.9 & 59.1 & 80.6 & 17.9 & 79.7 & 31.2  & 81.0 & 26.5  & 73.5 & 8.5  & 52.4   \\
			DANN \cite{ganin2016domain}&\xmark   & 81.9 & 77.7 & 82.8 & 44.3 & 81.2 & 29.5 & 65.1 & 28.6  & 51.9 & 54.6  & 82.8 & 7.8  & 57.4   \\
			DAN \cite{long2015learning}&\xmark   & 87.1 & 63.0 & 76.5 & 42.0 & 90.3 & 42.9 & 85.9 & 53.1  & 49.7 & 36.3  & 85.8 & 20.7 & 61.1   \\
			ADR \cite{saito2017adversarial}&\xmark& 94.2 & 48.5 & 84.0 & {72.9} & 90.1 & 74.2 & {92.6} & 72.5 & 80.8 & 61.8 & 82.2 & 28.8 & 73.5 \\
			CDAN \cite{long2018conditional}&\xmark  & 85.2 & 66.9 & 83.0 & 50.8 & 84.2 & 74.9 & 88.1 & 74.5  & 83.4 & 76.0  & 81.9 & 38.0 & 73.9   \\
			CDAN+BSP \cite{chen2019transferability}&\xmark & 92.4 & 61.0 & 81.0 & 57.5 & 89.0 & 80.6 & {90.1} & 77.0 & 84.2 & 77.9 & 82.1 & 38.4 & 75.9 \\
			SAFN \cite{Xu_2019_ICCV}&\xmark & 93.6 & 61.3 & {84.1} & 70.6 & {94.1} & 79.0 & 91.8 & {79.6} & {89.9} & 55.6 & {89.0} & 24.4 & 76.1 \\
			SWD \cite{lee2019sliced}&\xmark & 90.8 & {82.5} & 81.7 & 70.5 & 91.7 & 69.5 & 86.3 & 77.5 & 87.4 & 63.6 & 85.6 & 29.2 & 76.4 \\
			MDD \cite{zhang2019bridging}&\xmark& - & {-} & -& - & - & - & - & - & - & - & - & - & 74.6 \\
			DMRL \cite{wu2020dual}&\xmark& - & {-} & -& - & - & - & - & - & - & - & - & - & 75.5 \\
			DM-ADA~\cite{xu2019adversarial}&\xmark & - & {-} & -& - & - & - & - & - & - & - & - & - & 75.6 \\
			MCC~\cite{jin2019minimum}&\xmark& {88.7} & 80.3 & 80.5 & 71.5 & 90.1 & 93.2 & 85.0 & 71.6 & 89.4 & {73.8} & 85.0 & {36.9} & {78.8} \\
			STAR~\cite{lu2020stochastic}&\xmark& {95.0} & 84.0 & {84.6} & 73.0 & 91.6 & 91.8 & 85.9 & 78.4 & 94.4 & {84.7} & 87.0 & {42.2} & {82.7} \\
			RWOT~\cite{xu2020reliable}&\xmark& 95.1& 80.3& 83.7& \textbf{90.0}& 92.4& 68.0& \textbf{92.5}& 82.2& 87.9& 78.4& {90.4}& \textbf{68.2}& 84.0 \\
			RSDA-MSTN~\cite{gu2020spherical}&\xmark& - & {-} & -& - & - & - & - & - & - & - & - & - & 75.8 \\
			\toprule

			3C-GAN~\cite{li2020model}&\cmark & {94.8} & 73.4 & 68.8 & {74.8} & 93.1 & {95.4} & {88.6} & \textbf{84.7} & 89.1 & {84.7} & 83.5 & {48.1} & {81.6} \\
			
			SHOT~\cite{liang2020we}&\cmark & {94.3} & {88.5} & 80.1 & 57.3 & 93.1 & {94.9} & 80.7 & {80.3} & {91.5} & {89.1} & 86.3 & {58.2} & {82.9} \\
			
			HCL~\cite{huang2021model}&\cmark&93.3& 85.4& 80.7& 68.5& 91.0& 88.1& 86.0& 78.6& 86.6&88.8&80.0& {74.7}&83.5\\

			{\textbf{NRC}}&\cmark & {96.8} & {91.3} & {82.4} & 62.4 & {96.2} & {95.9} & 86.1 &{80.6} & {94.8} & {94.1} & 90.4  &  {59.7} & {85.9} \\
			{\textbf{NRC++}}&\cmark & \textbf{97.4} & \textbf{91.9} & \textbf{88.2} & 83.2 & \textbf{97.3} & \textbf{96.2} & 90.2 &{81.1} & \textbf{96.3} & \textbf{94.3} & \textbf{91.4}  &  {49.6} & \textbf{88.1} \\
			\hline
			\end{tabular}}
			
			\label{tab:visda}
	\end{center}
\end{table*}

\begin{table*}[t]
\begin{center}
\caption{Accuracies (\%) on PointDA-10. \textit{The results except ours are from PointDAN~\cite{qin2019pointdan}}.}\label{tab:point}
\scalebox{0.99}{
\setlength{\tabcolsep}{4mm}{ 
  \begin{tabular}{l|c|ccc cc cc}
\toprule 
\multicolumn{1}{c}{}&\multicolumn{1}{|c|}{SF} &\multicolumn{1}{c}{Mo$\rightarrow$Sh}  &\multicolumn{1}{c}{Mo$\rightarrow$Sc}&\multicolumn{1}{c}{Sh$\rightarrow$Mo}&\multicolumn{1}{c}{Sh$\rightarrow$Sc}&\multicolumn{1}{c}{Sc$\rightarrow$Mo}&\multicolumn{1}{c}{Sc$\rightarrow$Sh}&\multicolumn{1}{c}{Avg}\\
\toprule
\multicolumn{1}{c}{MMD~\cite{long2013transfer}}&\multicolumn{1}{|c|}{\xmark} &\multicolumn{1}{c}{57.5} &\multicolumn{1}{c}{27.9} &\multicolumn{1}{c}{40.7}  &\multicolumn{1}{c}{26.7} &\multicolumn{1}{c}{47.3} &\multicolumn{1}{c}{54.8}  &\multicolumn{1}{c}{42.5}\\
\multicolumn{1}{c}{DANN~\cite{ganin2014unsupervised}}&\multicolumn{1}{|c|}{\xmark} &\multicolumn{1}{c}{58.7} &\multicolumn{1}{c}{29.4} &\multicolumn{1}{c}{42.3}  &\multicolumn{1}{c}{30.5} &\multicolumn{1}{c}{48.1} &\multicolumn{1}{c}{56.7}  &\multicolumn{1}{c}{44.2}\\
\multicolumn{1}{c}{ADDA~\cite{tzeng2017adversarial}}&\multicolumn{1}{|c|}{\xmark} &\multicolumn{1}{c}{61.0} &\multicolumn{1}{c}{30.5} &\multicolumn{1}{c}{40.4}  &\multicolumn{1}{c}{29.3} &\multicolumn{1}{c}{48.9} &\multicolumn{1}{c}{51.1}  &\multicolumn{1}{c}{43.5}\\
\multicolumn{1}{c}{MCD~\cite{saito2018maximum}}&\multicolumn{1}{|c|}{\xmark} &\multicolumn{1}{c}{62.0} &\multicolumn{1}{c}{31.0} &\multicolumn{1}{c}{41.4}  &\multicolumn{1}{c}{31.3} &\multicolumn{1}{c}{46.8} &\multicolumn{1}{c}{59.3}  &\multicolumn{1}{c}{45.3}\\

\multicolumn{1}{c}{PointDAN~\cite{qin2019pointdan}}&\multicolumn{1}{|c|}{\xmark} &\multicolumn{1}{c}{{64.2}} &\multicolumn{1}{c}{\textbf{33.0}} &\multicolumn{1}{c}{{47.6}}  &\multicolumn{1}{c}{\textbf{33.9}} &\multicolumn{1}{c}{{49.1}} &\multicolumn{1}{c}{{64.1}}  &\multicolumn{1}{c}{{48.7}}\\

\toprule
\multicolumn{1}{c}{Source-only}&\multicolumn{1}{|c|}{} &\multicolumn{1}{c}{43.1} &\multicolumn{1}{c}{17.3} &\multicolumn{1}{c}{40.0}  &\multicolumn{1}{c}{15.0} &\multicolumn{1}{c}{33.9} &\multicolumn{1}{c}{47.1}  &\multicolumn{1}{c}{32.7}\\


\multicolumn{1}{c}{{\textbf{NRC}}}&\multicolumn{1}{|c|}{\cmark}&\multicolumn{1}{c}{{64.8}} &\multicolumn{1}{c}{{25.8}} &\multicolumn{1}{c}{{59.8}}  &\multicolumn{1}{c}{{26.9}} &\multicolumn{1}{c}{{70.1}} &\multicolumn{1}{c}{{68.1}}  &\multicolumn{1}{c}{{52.6}}\\

\multicolumn{1}{c}{{\textbf{NRC++}}}&\multicolumn{1}{|c|}{\cmark}&\multicolumn{1}{c}{\textbf{67.2}} &\multicolumn{1}{c}{{27.6}} &\multicolumn{1}{c}{\textbf{60.2}}  &\multicolumn{1}{c}{{30.4}} &\multicolumn{1}{c}{\textbf{74.5}} &\multicolumn{1}{c}{\textbf{71.2}}  &\multicolumn{1}{c}{\textbf{55.1}}\\
\hline
\end{tabular}
}}
\end{center}
\end{table*}

\subsection{Vanilla Domain Adaptation}
\noindent \textbf{2D image datasets.} We first evaluate the target performance of our method compared with existing DA and SFDA methods on three 2D image datasets. As shown in Tab.~\ref{tab:office31}-\ref{tab:visda}, the top part shows results for the source-present methods \textit{with access to source data during adaptation}. The bottom shows results for the source-free DA methods. 
On Office-31, our method gets similar results compared with source-free method 3C-GAN and lower than source-present method RSDA. And our method achieves state-of-the-art performance on Office-Home and VisDA, especially on VisDA our method surpasses the source-free method SHOT and source-present method RWOT by a wide margin (3\% and 1.9\% respectively). {When {excluding potential outliers}, as done by our method (i.e., \textit{NRC++} ), we outperform all baselines and NRC. Especially for the VisDA dataset, we improve the accuracy from 85.9$\%$ to 88.1$\%$. } 
The reported results clearly demonstrate the efficiency of the proposed method for source-free domain adaptation. Interestingly, like already observed in the SHOT paper, source-free methods outperform methods that have access to source data during adaptation. 

\noindent \textbf{3D point cloud dataset.} We also report the result for the PointDA-10. As shown in Tab.~\ref{tab:point}, our method outperforms  PointDA~\cite{qin2019pointdan}, which demands source data for adaptation and is specifically tailored for point cloud data with extra attention modules, by a large margin (4\%).{ Similarly, we can draw the same conclusion: introducing the density loss helps us to reduce the negative impact of the outliers resulting in better performance for \textit{NRC++}.} 

\subsection{Partial-set domain adaptation}
We also show that our method can be extended to partial-set domain adaptation, where the target label space is a subset of the source domain. {The challenge here is that the model may fail to distinguish which categories the target samples come from.
Specially for the datatset we use, \textit{i.e.}, Office-Home,} there are totally 25 classes (the first 25 in the alphabetical order) out of 65 classes in the target domain for \textbf{Office-Home} (as also used in \cite{liang2020we}). {Here we directly deploy our method to source-free partial-set DA without introducing extra processes}. As reported in Tab.~\ref{tab:pda}, our \textbf{NRC} has better result than source-aware methods, and slightly outperforms SHOT. \textbf{NRC++} does not lead to a large performance gain on this setting. {The results indicate the generalization ability of our method.} 


\begin{table*}[btp]
\caption{Ablation study of different modules on Office-Home (\textbf{left}) and VisDA (\textbf{middle}), comparison between using expanded neighbors and larger nearest neighbors (\textbf{right}).}\label{tab:Ablation}
\begin{minipage}[t]{.35\linewidth}
	\makeatletter 
    {\setlength{\tabcolsep}{6pt}\renewcommand{\arraystretch}{.7}
\begin{tabular}{ccccccc|c}
					\hline
					$\mathcal{L}_{div}$&$\mathcal{L}_{\mathcal{N}}$&$\mathcal{L}_E$&$\mathcal{L}_{\bm{\hat{E}}}$&A&$\mathcal{L}_D$& \multicolumn{1}{|c}{Avg} \\
					\hline
					& & & &&& \multicolumn{1}{|c}{59.5}\\
					\bm{\cmark}& & && && \multicolumn{1}{|c}{62.1}\\
					\bm{\cmark}&\bm{\cmark}& && && \multicolumn{1}{|c}{69.1}\\
					\bm{\cmark}&\bm{\cmark}&  &&\bm{\cmark}  &&\multicolumn{1}{|c}{71.1}\\
					\bm{\cmark}&\bm{\cmark}& \bm{\cmark}&   &&  &\multicolumn{1}{|c}{65.2}\\
					\bm{\cmark}&\bm{\cmark}& \bm{\cmark} && \bm{\cmark}&& \multicolumn{1}{|c}{72.2}\\
					\bm{\cmark}&\bm{\cmark}& \bm{\cmark} && \bm{\cmark}&\bm{\cmark}& \multicolumn{1}{|c}{\textbf{72.5}}\\
					\bm{\cmark}&\bm{\cmark}&  &\bm{\cmark}& \bm{\cmark}& &\multicolumn{1}{|c}{{69.1}}\\
					\hline
		\end{tabular}}
\end{minipage}
\begin{minipage}[t]{.35\linewidth}
	\makeatletter 
    {\setlength{\tabcolsep}{6pt}\renewcommand{\arraystretch}{.7}
\begin{tabular}{cccccc|c}
					\hline					$\mathcal{L}_{div}$&$\mathcal{L}_{\mathcal{N}}$&$\mathcal{L}_E$&$\mathcal{L}_{\bm{\hat{E}}}$&A&$\mathcal{L}_D$& \multicolumn{1}{|c}{Acc} \\
					\hline
					&& & &&& \multicolumn{1}{|c}{44.6}\\
					\bm{\cmark}&& &  &&&\multicolumn{1}{|c}{47.8}\\
					\bm{\cmark}&\bm{\cmark}&  &&  &&\multicolumn{1}{|c}{81.5}\\
					\bm{\cmark}&\bm{\cmark}&  &&\bm{\cmark} &&\multicolumn{1}{|c}{82.7}\\
					\bm{\cmark}&\bm{\cmark}& \bm{\cmark} &    &&&\multicolumn{1}{|c}{61.2}\\
					\bm{\cmark}&\bm{\cmark} &\bm{\cmark}& & \bm{\cmark} & & \multicolumn{1}{|c}{85.9}\\
					\bm{\cmark}&\bm{\cmark} &\bm{\cmark}& & \bm{\cmark} & \bm{\cmark}& \multicolumn{1}{|c}{\textbf{88.1}}\\
					\bm{\cmark}&\bm{\cmark}& &\bm{\cmark} & \bm{\cmark} & & \multicolumn{1}{|c}{{82.0}}\\
					\hline
		\end{tabular}}
	\end{minipage}
\begin{minipage}[t]{.2\linewidth}
	\makeatletter 
    {\setlength{\tabcolsep}{6pt}\renewcommand{\arraystretch}{1.5}
\begin{tabular}{c|c}
					\hline
					\textbf{Method\&Dataset}& \multicolumn{1}{|c}{Acc} \\
					\hline
					\text{VisDA} ($K$=$M$=5)& \multicolumn{1}{|c}{\textbf{85.9}}\\
					\text{VisDA} w/o $E$ ($K$=30)& \multicolumn{1}{|c}{{84.0}}\\
					\hline
					\text{OH} ($K$=3,$M$=2)& \multicolumn{1}{|c}{\textbf{72.2}}\\
					\text{OH} w/o $E$ ($K$=9)& \multicolumn{1}{|c}{{69.5}}\\
					\hline
		\end{tabular}}
	\end{minipage}
\end{table*}

\begin{table}[t]
\caption{{Runtime analysis on SHOT and our method. For SHOT, pseudo labels are computed at each epoch. 20\%, 10\% and 5\% denote the percentage of target features which are stored in the memory bank.}}\label{tab:time}
	\begin{center}
	\makeatletter 
		\linespread{3.0}
		\setlength{\tabcolsep}{2mm}
\begin{tabular}{cc|c}
					\hline
					VisDA& Runtime (s/epoch) & Per-class (\%) \\
					\hline
				    SHOT&618.82&82.9\\
				    \hline
				    \textbf{NRC}&540.89&85.9\\
				    \textbf{NRC}(20\% for memory bank)&507.15&85.3\\
				    \textbf{NRC}(10\% for memory bank)&499.49&85.2\\
				    \textbf{NRC}(5\% for memory bank)&499.28&85.1\\
					\hline
		\end{tabular}
	\end{center}
	\end{table}

\begin{figure*}[t]
	\centering
	\includegraphics[width=0.99\textwidth]{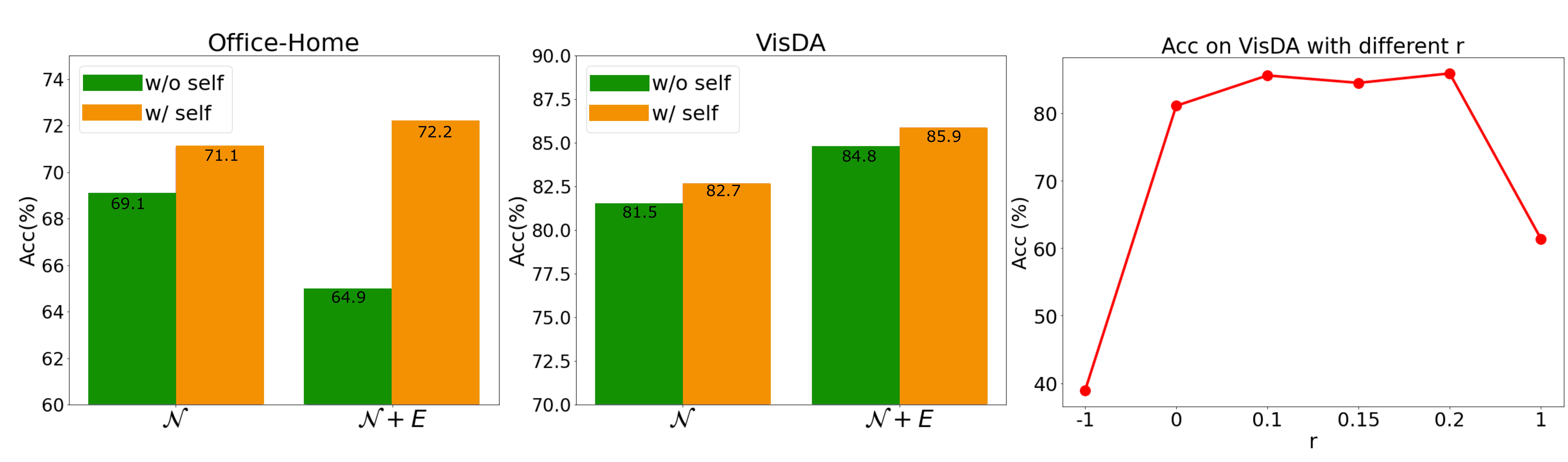}
	\caption{(\textbf{Left and middle}) Ablation study of $\mathcal{L}_{self}$ on Office-Home and VisDA respectively. (\textbf{Right}) Performance with different $r$ on VisDA. 
	}
	\label{fig:abas}
\end{figure*}

\begin{table}[t]
\caption{\SQ{Analysis of used prior information for the target category distribution in the diversity loss $\mathcal{L}_{div}$, on Ar$\rightarrow$Cl, Office-Home.}}\label{tab:div}
	\begin{center}
	\makeatletter 
		\linespread{3.0}
		\setlength{\tabcolsep}{2mm}
\begin{tabular}{cc}
					\hline
					Prior information & Per-class (\%) \\
					\hline
				    Uniform distribution & 57.7\\
                        Real target category distribution & 56.9\\
					\hline
		\end{tabular}
	\end{center}
	\end{table}



\subsection{Multi-Source Domain Adaptation}
{
We also evaluate our method on the multi-source single-target setting on Office-Home and the large-scale DomainNet benchmark. {The difference between single-source (normal) domain adaptation and multi-source domain adaptation is that in multi-source DA the domain shift between each source domain may deteriorate the model training. However, here we directly deploy our method to source-free multi-source domain adaptation, where the training stages are similar to source-free domain adaptation, except that the source model is trained with data from multiple source domains.} \SQ{The SHOT methods in Tab.~\ref{tab:msda_compare} are the baselines for source-free multi-source DA, where SHOT w/o domain labels means only using one source model, while SHOT-Ens (the reported results are from DECISION~\cite{DECISION}) means using multiple source models, their results indicate that using multiple source model could further improve the performance. As reported in Tab.~\ref{tab:msda_compare}, without using domain labels, we are able to achieve the best score on the challenging DomainNet benchmark compared to the source-free multi-source DA methods}, and comparable ones with baselines on office-home. For example, comparing with SHOT-Ens,  on the DomainNet dataset we observe improvements of ${1.1\%}$  despite not using domain labels. NRC++ further improves performance from 47.3\% to 48.2\%.}
 
\subsection{Multi-Target domain adaptation}
{
We also evaluate our method for single-source multi-target domain adaptation on Office-31. {In multi-target domain adaptation, the model is trained with a single labeled source domain and multiple unlabeled target domains, the final goal is to learn a good classifier for all target domains. Like multi-source domain adaptation, directly deploying a normal domain adaptation method to multi-target domain adaptation will usually lead to bad performance, due to the negative transfer~\cite{chen2019blending} caused by the different target domains. In this subsection, we show that our method can directly work quite well under multi-target domain adaptation, even under the source-free setting.} As reported in Tab.~\ref{tab:mtda_office-31},  the source model has the worst result (i.e., 68.4~$\%$) without any domain adaptation technique. Using both the source data and the domain label, D-CGCT achieve the best score (i.e., 88.8~$\%$). While our method, without both the source data and the domain label, still obtains 85.0~$\%$ accuracy, which indicates that our method gets comparable results even under this more challenging setting. 

}
 
\begin{table*}[t]
    \centering
    \caption{{Accuracy on both DomainNet and Office-Home for Multi-Source Domain Adaptation.}}
    \label{tab:msda_compare}
    \setlength{\tabcolsep}{5pt}
    \resizebox{1\linewidth}{!}{%
        \begin{tabular}{lcccccccclcccccl}
            \toprule
            \multirow{2}{30pt}{\centering Method} & \multirow{2}{*}{\centering SF} & \multirow{2}{*}{\parbox{3cm}{\centering   w/o Domain \\ Labels}} & \multicolumn{7}{c}{\textbf{DomainNet}
            } && \multicolumn{5}{c}{\textbf{Office-Home}} \\
            \cmidrule(lr){4-10} \cmidrule(lr){12-16}
            & & & $\mapsto $C &  $\mapsto $I &  $\mapsto $P &  $\mapsto $Q &  $\mapsto $R &  $\mapsto $S & Avg & &  $\mapsto $Ar &  $\mapsto $Cl &  $\mapsto $Pr & $\mapsto $Rw & Avg \\
            \midrule
            {SImpAl$_{50}$~\cite{SImpAl}} & \xmark & \xmark &   66.4 & 26.5 & 56.6 & 18.9 & 68.0 & 55.5 & 48.6 & & 70.8 & 56.3 & 80.2 & 81.5 & 72.2 \\ 
            {CMSDA~\cite{CMSDA}} & \xmark & \xmark &  70.9 & 26.5 & 57.5 & 21.3 & 68.1 & 59.4 & 50.4 & &  71.5 & \textbf{67.7} & 84.1 & 82.9 & \textbf{76.6} \\ 
            {DRT \cite{DRT}} & \xmark & \xmark &   71.0 & \textbf{31.6} & 61.0 & 12.3 & 71.4 & \textbf{60.7} & 51.3 &  & - & - & - & - & -\\ 
            STEM~\cite{STEM} & \xmark & \xmark & \textbf{72.0} & 28.2 & \textbf{61.5} & \textbf{25.7} & \textbf{72.6} & 60.2 & \textbf{53.4} & & - & - & - & - & -  \\ 
            \midrule
            DECISION~\cite{DECISION} & \cmark & \xmark &61.5	 &21.6	 &54.6 	 & {18.9}	 &67.5	&51.0	 &45.9 & & {74.5} &{59.4} &{84.4} & {83.6} &{75.5}\\
            CAiDA~\cite{caida} &\cmark & \xmark & - & - & - & - & - & - & - & & \textbf{75.2} & 60.5 & {84.7} & \textbf{84.2} & 76.2\\
            \SQ{SHOT~\cite{liang2020we}} & \cmark & \cmark& {58.3}& 22.7& 53.0 &18.7 &65.9& 48.4& 44.5& & 72.1& 57.2& 83.4& 81.3& 73.5\\
            SHOT~\cite{liang2020we}-Ens & \cmark & \xmark &58.6	&25.2	&55.3	&15.3	&70.5	&52.4	&46.2 & & 72.2 &59.3 &82.8 &82.9 &74.3\\
            \midrule
            {Source} &\xmark & \cmark &57.0 &23.4 &54.1 &14.6 &67.2 &50.3 & 44.4 & & 58.0 & 57.3 & 74.2 & 77.9 & 66.9\\
            \textbf{NRC} & \cmark & \cmark &  65.3 & 24.4 & 55.9 & 16.1 & 69.3 & 53.0 & 47.3 & & 70.8 & 60.1 & 84.8 & 83.7 & 74.8 \\
            \textbf{NRC++} & \cmark & \cmark & 66.1 & 24.8 & 57.2 & 17.3 & 70.1 & 54.0 & 48.2 & & 71.2 & 61.1 & \textbf{84.9 }& 83.8 & 75.3 \\            
            \bottomrule
            \end{tabular} 
        }
\end{table*}

\begin{table*}[t]
    \centering
    \caption{{Accuracy on Office-31 for Multi-Target Domain Adaptation. * indicates taken from CGCT~\cite{CGCT}}}
    \label{tab:mtda_office-31}
    \setlength{\tabcolsep}{15pt}
    \resizebox{1\linewidth}{!}{%
        \begin{tabular} {lcccccc}
        \toprule
        \multirow{2}{30pt}{\centering Method} & \multirow{2}{*}{\centering SF} & \multirow{2}{*}{\parbox{3cm}{\centering   w/o Domain \\ Labels}} & \multicolumn{4}{c}{\textbf{Office-31}}\\
        \cmidrule{4-7}
        & && Amazon$\mapsto $ & DSLR$\mapsto $ & Webcam$\mapsto $& Avg. \\
        \midrule
        Source model & \xmark & \cmark & 68.6 & 70.0 & 66.5 & 68.4 \\
        \midrule
        MT-MTDA~\cite{nguyen2020unsupervised} & \xmark & \xmark & 87.9 & 83.7 & 84.0 & 85.2 \\
        HGAN~\cite{yang2020heterogeneous} & \xmark & \xmark & 88.0 & 84.4 & 84.9 & 85.8 \\
        D-CGCT~\cite{CGCT} & \xmark & \xmark & {93.4} & \textbf{86.0} & \textbf{87.1} & \textbf{88.8} \\
        \midrule 
        JAN~\cite{long2017deep}* & \xmark & \cmark & 84.2 & 74.4 & 72.0 & 76.9 \\
        CDAN~\cite{long2018conditional}* & \xmark & \cmark & 93.6	& 80.5	& 81.3	& 85.1 \\
        AMEAN~\cite{chen2019blending} & \xmark & \cmark & 90.1 & 77.0 & 73.4 & 80.2 \\
        GDA~\cite{GDA}  & \xmark & \cmark & 88.8 & 74.5 & 73.2 & 87.9\\
        CGCT~\cite{CGCT}  & \xmark & \cmark & 93.9 & 85.1 & 85.6 & 88.2\\
        \hline
        \textbf{NRC}  & \cmark  & \cmark & 93.5 & 80.1 & 79.3 & 84.3\\
        \textbf{NRC++}  & \cmark  & \cmark & \textbf{95.3} & 80.2 & 79.5 & 85.0\\
        \bottomrule
        \end{tabular}
        }
     
\end{table*}


\begin{table*}[tbp]
    \vspace{-0mm}
    \centering
    \caption{{Accuracy on Office-Home using ResNet-50 as backbone for {\textbf{partial-set DA}}}. 
    \vspace{-0mm}}
    \label{tab:pda}
    \setlength{\tabcolsep}{1pt}
    \resizebox{0.95\textwidth}{!}{%
        \begin{tabular}{cccccccccccccccccc}
            \hline
            {\textbf{Partial-set}}& \textbf{SF}  & Ar$\rightarrow$Cl & Ar$\rightarrow$Pr & Ar$\rightarrow$Re & Cl$\rightarrow$Ar & Cl$\rightarrow$Pr & Cl$\rightarrow$Re & Pr$\rightarrow$Ar & Pr$\rightarrow$Cl & Pr$\rightarrow$Re & Re$\rightarrow$Ar & Re$\rightarrow$Cl & Re$\rightarrow$Pr & \textbf{Avg}. \\
				\hline
			    ResNet-50 \cite{he2016deep}& \xmark & 46.3 & 67.5 & 75.9 & 59.1 & 59.9 & 62.7 & 58.2 & 41.8 & 74.9 & 67.4 & 48.2 & 74.2 & 61.3\\
			    IWAN \cite{zhang2018importance}& \xmark & 53.9 & 54.5 & 78.1 & 61.3 & 48.0 & 63.3 & 54.2 & 52.0 & 81.3 & 76.5 & 56.8 & 82.9 & 63.6  \\
			    SAN \cite{cao2018partiala}& \xmark & 44.4 & 68.7 & 74.6 & 67.5 & 65.0 & 77.8 & 59.8 & 44.7 & 80.1 & 72.2 & 50.2 & 78.7 & 65.3  \\
			    DRCN~\cite{li2020deep_pda}& \xmark & 54.0 & 76.4 & 83.0 & 62.1 & 64.5 & 71.0 & 70.8 & 49.8 & 80.5 & 77.5 & 59.1 & 79.9 & 69.0  \\
			    ETN \cite{cao2019learning}& \xmark & 59.2 & 77.0 & 79.5 & 62.9 & 65.7 & 75.0 & 68.3 & 55.4 & 84.4 & 75.7 & 57.7 & 84.5 & 70.5 \\
			    SAFN \cite{xu2019larger}& \xmark & 58.9 & 76.3 & 81.4 & 70.4 & 73.0 & 77.8 & 72.4 & 55.3 & 80.4 & 75.8 & 60.4 & 79.9 & 71.8 \\
			    RTNet$_{adv}$~\cite{chen2020selective}& \xmark & 63.2 & 80.1 & 80.7 & 66.7 & 69.3 & 77.2 & 71.6 & 53.9 & 84.6 & 77.4 & 57.9 & 85.5& 72.3 \\
			    BA$^3$US~\cite{liang2020balanced}& \xmark & 60.6 & 83.2 & 88.4 & 71.8 & 72.8 & 83.4 & 75.5 & 61.6 & 86.5 & 79.3 & 62.8 & 86.1 & 76.0 \\
        	    TSCDA~\cite{ren2020learning}& \xmark & 63.6 & 82.5 & 89.6 & 73.7 & 73.9 & 81.4 & 75.4 & 61.6 & 87.9 & \textbf{83.6} & 67.2 & 88.8 & 77.4 \\
            \hline
				SHOT-IM~\cite{liang2020we}& \cmark  & 57.9 & 83.6 & 88.8 & 72.4 & 74.0 & 79.0 & 76.1 & 60.6 & {90.1} & {81.9} & {\textbf{68.3}} & 88.5 & 76.8 \\
				SHOT~\cite{liang2020we}& \cmark  & {64.8} & {\textbf{85.2}} & {92.7} & {76.3} & {\textbf{77.6}} & {\textbf{88.8}} & {\textbf{79.7}} & {64.3} & 89.5 & 80.6 & {66.4} & 85.8 & {79.3} \\
				\textbf{NRC}& \cmark & 66.2 & 84.2 &  \textbf{92.9} & 77.5  & 75.2   & 83.1 & 76.6  & 68.1  & {88.3}  &82.4   & 67.5 &  \textbf{88.6}  & 79.5  \\
   				\textbf{NRC++}& \cmark  & {\textbf{66.3}}   & 85.0 &  92.8 & \textbf{78.0}  & 75.3   & 83.5 & 76.7  & \textbf{68.3}  & {\textbf{90.6}}  &82.5   & 67.7 &  88.5  & \textbf{79.6}  \\        
            \hline
        \end{tabular}
    }
\end{table*}


\subsection{Analysis}

\noindent \textbf{Ablation study on neighbors $\mathcal{N}$, $E$,  affinity $A$ and {density loss $\mathcal{D}$}.} In the first two tables of Tab.~\ref{tab:Ablation}, we conduct the ablation study on Office-Home and VisDA. The 1-st row contains results from the source model and the 2-nd row from only training with the diversity loss $\mathcal{L}_{div}$. From the remaining rows, several conclusions can be drawn.

First, the original supervision, which considers all neighbors equally can lead to a decent performance (69.1 on Office-Home). 
Second, considering higher affinity values for reciprocal neighbors leads to a large performance gain (71.1 on Office-Home). Last but not the least, the expanded neighborhoods can also be helpful, but only when combined with the affinity values $A$ (72.2 on Office-Home). 
Using expanded neighborhoods without affinity obtains bad performance (65,2 on Office-Home). We conjecture that those expanded neighborhoods, especially those neighbors of nRNN, may be noisy as discussed in Sec.~\ref{sec:expand}. Removing the affinity $A$ means we treat all those neighbors equally, which is not reasonable. {Furthermore, as reported in the penultimate rows (Tab.~\ref{tab:Ablation}(left, middle)) outlier exclusion (with $\mathcal{L}_D$) further improves the model performance (e.g., from 85.9 to 88.1 on VisDA), indicating that considering the density around each samples is useful and empirically effective.}

We also show that duplication in the expanded neighbors is important in the last  row of Tab.~\ref{tab:Ablation}, where the $\mathcal{L}_{\bm{\hat{E}}}$ means we remove duplication in Eq.~\ref{eq:nn2}. The results show that the performance will degrade significantly when removing them, implying that the duplicated expanded neighbors are indeed more important than others.

Next we ablate the importance of the expanded neighborhood in the right of Tab. \ref{tab:Ablation}. We show that if we increase the number of datapoints considered for class-consistency by simply considering a larger K, we obtain significantly lower scores. We have chosen $K$ so that the total number of points considered is equal to our method (i.e. 5+5*5=30 and 3+3*2=9).
Considering neighbors of neighbors is more likely to provide datapoints that are close on the data manifold~\cite{tenenbaum2000global}, and are therefore more likely to share the class label with the ego feature. 

\begin{figure}[t]
	\centering
	\includegraphics[height=0.15\textheight,width=\columnwidth]{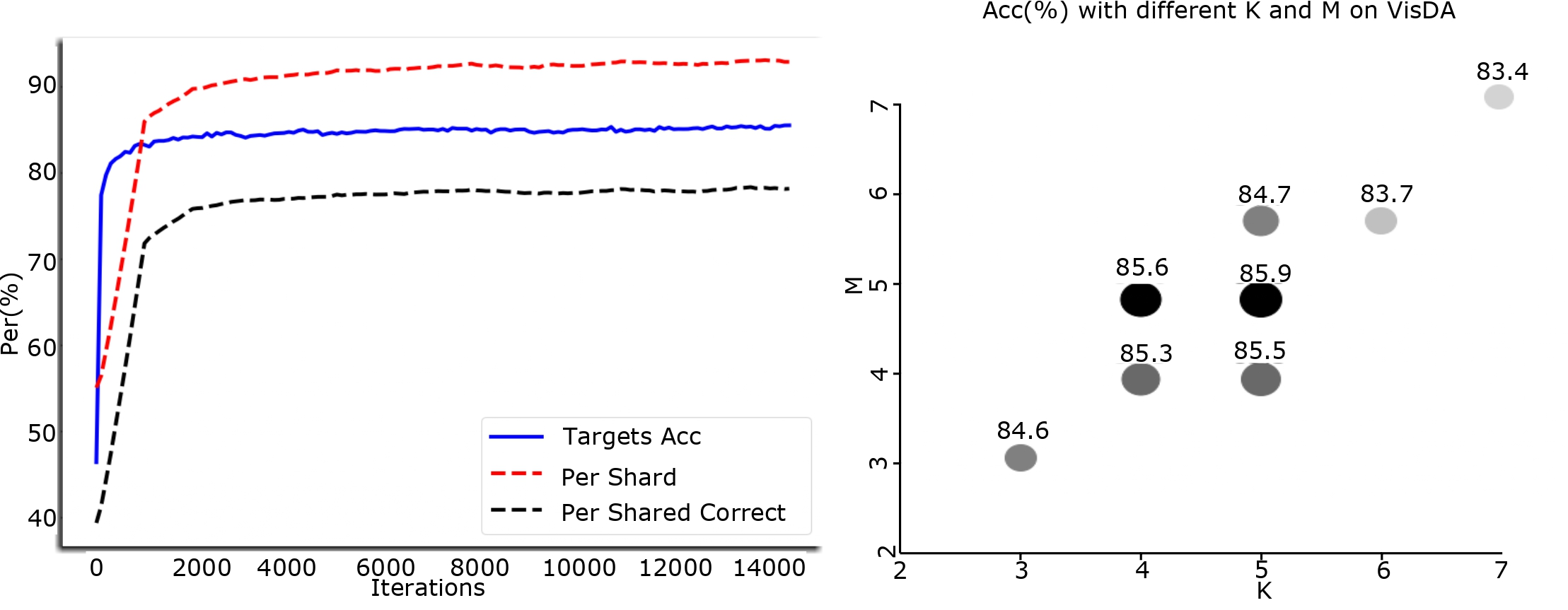}
	\caption{(\textbf{Left}) The three curves are (on VisDA): target accuracy (\textit{Blue}), ratio of features which have 5-nearest neighbors all sharing the same predicted label (\textit{dashed Red}), and ratio of features which have 5-nearest neighbors all sharing the same and \textit{correct} predicted label (\textit{dashed Black}). (\textbf{Right}) Ablation study on choice of K and M on VisDA. 
	}
	\label{fig:curve_km}
\end{figure}

\begin{figure}[t]
	\centering
	\includegraphics[width=\columnwidth]{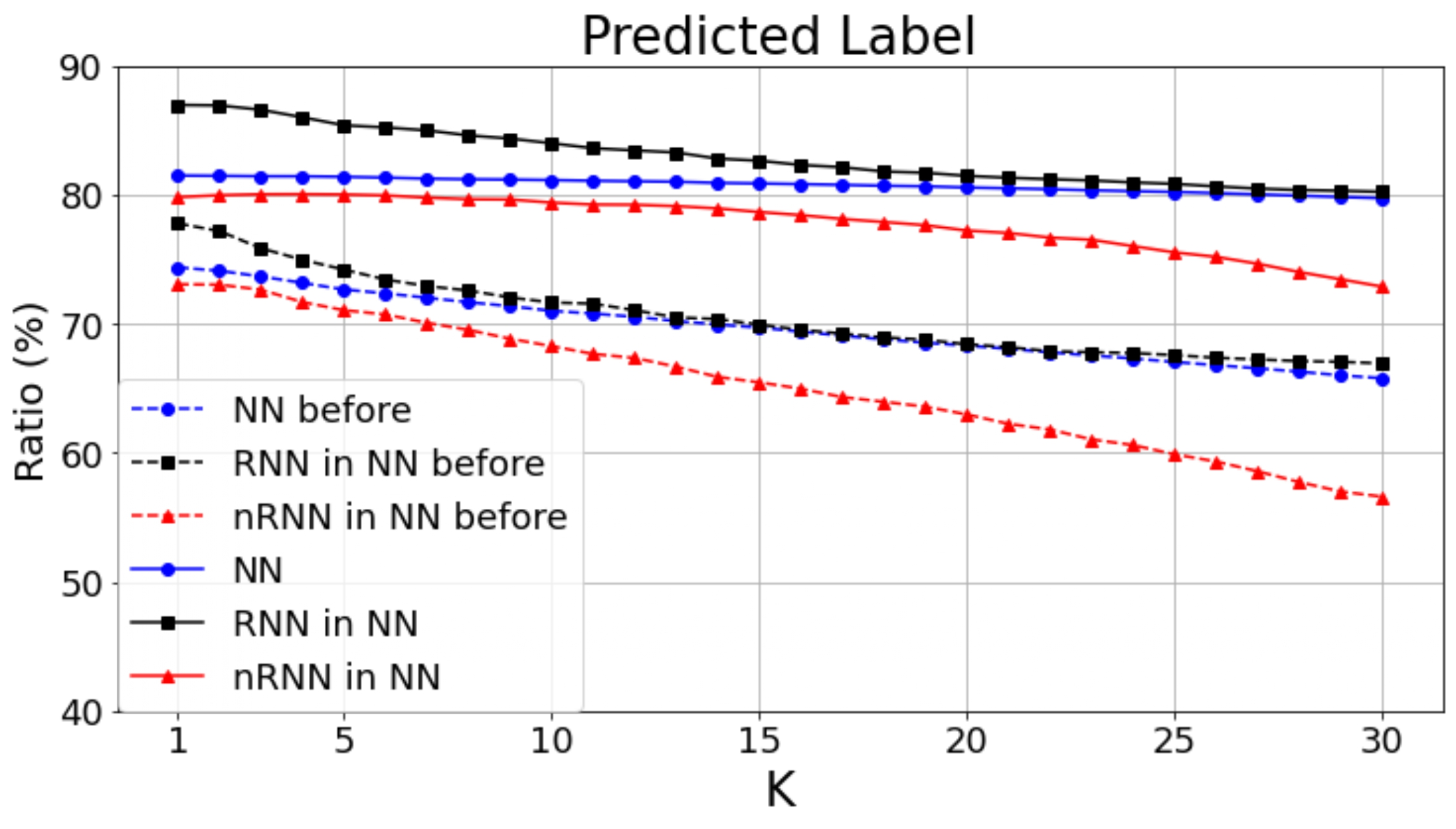}
	\caption{(Ratio of different type of nearest neighbor features which have the correct predicted label, before and after adaptation. }
	\label{fig:curves_sne}
\end{figure}

{\noindent \textbf{Runtime analysis.}  Instead of storing all feature vectors in the memory bank, we follow the same memory bank setting as in \cite{dwibedi2021little} which is for nearest neighbor retrieval. The method only stores a fixed number of target features, we update the memory bank at the end of each iteration by taking the $n$ (batch size) embeddings from the current training iteration and concatenating them at the end of the memory bank, and discard the oldest $n$ elements from the memory bank. We report the results with this type of memory bank of different buffer size in the Tab.~\ref{tab:time}. The results show that indeed this could be an efficient way to reduce computation on very large datasets.}

\SQ{\noindent \textbf{Analysis on the prior information for the target category distribution in $\mathcal{L}_{div}$.} In Tab.~\ref{tab:div}, we show the different choice of the prior information for the target category distribution in $\mathcal{L}_{div}$. Originally we use the uniform distribution, here we also use the ground truth target class distribution. The results show simply utilizing uniform distribution is enough, even surpassing the one with the real class distribution. We posit that the reason may be due to the mini-batch training, as every mini-batch may have different label distribution.}

\noindent \textbf{Ablation study on self-regularization.} In the left and middle of Fig~\ref{fig:abas}, we show the results with and without self-regularization $\mathcal{L}_{self}$. The $\mathcal{L}_{self}$ can improve the performance when adopting only nearest neighbors $\mathcal{N}$ or all neighbors $\mathcal{N}+E$. The results imply that self-regularization can effectively reduce the negative impact of the potential noisy neighbors, especially on the Office-Home dataset.

\noindent \textbf{Sensitivity to hyperparameter.} There are three hyperparameters in our method: K and M which are the number of nearest neighbors and expanded neighbors, $r$ which is the affinity value assigned to nRNN. We show the results with different $r$ in the right of Fig.~\ref{fig:abas}. \textit{Note we keep the affinity of expanded neighbors as 0.1}. $r=1$ means no affinity. $r=-1$ means treating supervision of nRNN feature as totally wrong, which is not always the case and will lead to quite lower result. $r=0$ can also achieve good performance, signifying RNN can already work well. Results with $r=0.1/0.15/0.2$ show that our method is not sensitive to the choice of a reasonable $r$.
{Note in DA, there is no validation set for hyperparameter tuning, we show the results varying the number of neighbors in the right of \SQ{Fig}.~\ref{fig:curve_km}, demonstrating the robustness to the choice of $K$ and $M$.}

\noindent \textbf{Training curve.} We show the evolution of several statistics during adaptation on VisDA in the left of \SQ{Fig}.~\ref{fig:curve_km}. The blue curve is the target accuracy. The dashed red and black curves are the ratio of features which have 5-nearest neighbors all sharing the same (\textit{dashed Red}), or the same and also \textbf{correct} (\textit{dashed Black}) predicted label. The curves show that the target features are clustering during the training. {Another interesting finding is that the curve 'Per Shared' correlates with the accuracy curve, which might therefore be used to determine training convergence.}

\noindent \textbf{Accuracy of supervision from neighbors.} We also show the accuracy of supervision from neighbors on task Ar$\rightarrow$Rw of Office-Home
in Fig.~\ref{fig:curves_sne}. 
It shows that after adaptation, the ratio of all types of neighbors having more correct predicted label, proving the effectiveness of the method.

\begin{figure}[tbp]
	\centering
	\includegraphics[width=\columnwidth]{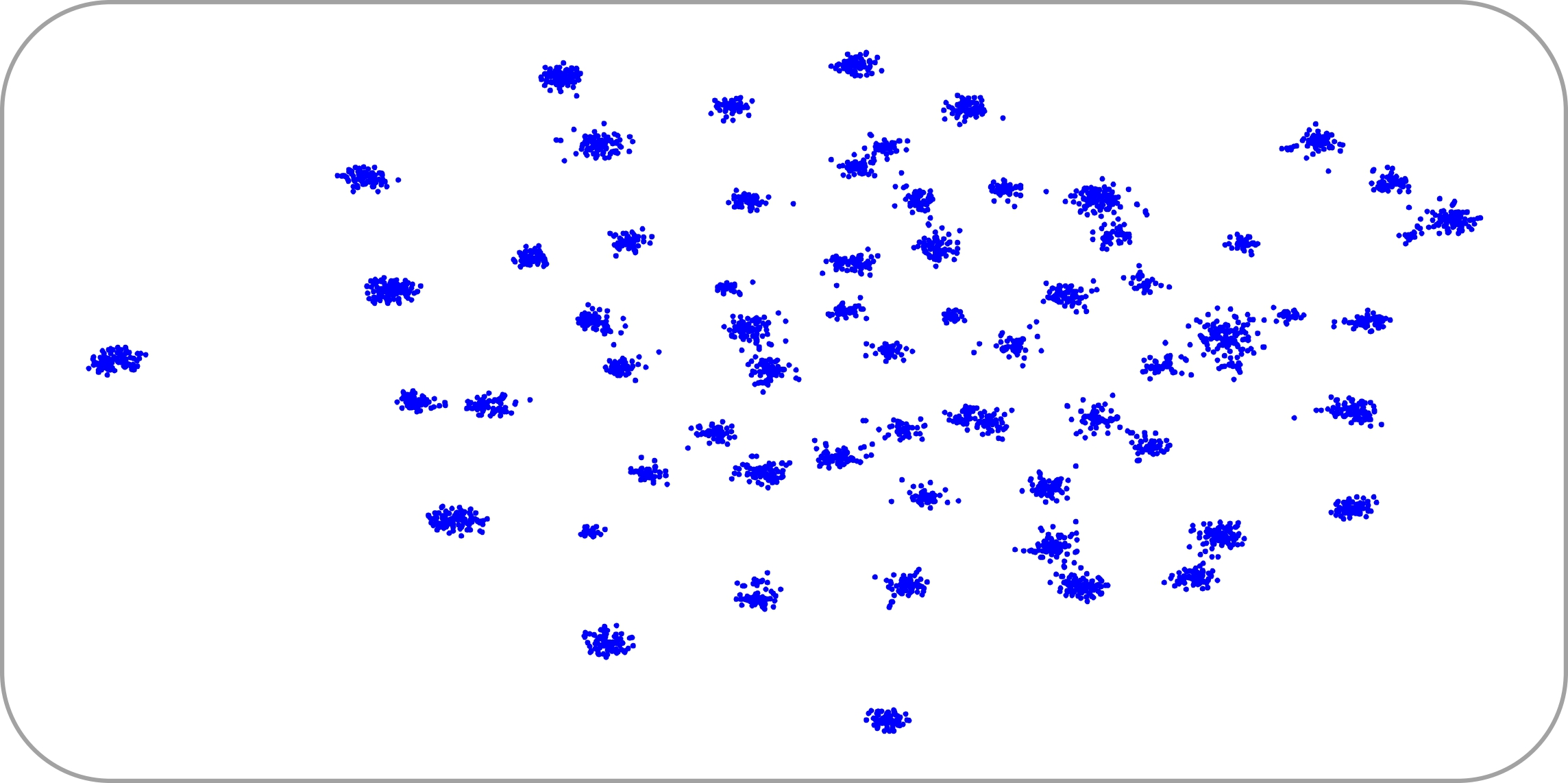}
	\caption{ Visualization of target features after adaptation.}
	\label{fig:sne}
\end{figure}

\noindent \textbf{t-SNE visualization.} We show the t-SNE feature visualization on task Ar$\rightarrow$Rw of target features before (Fig.~\ref{fig:motivation}(c)) and after (Fig.~\ref{fig:sne}) adaptation. After adaptation, the features are more compactly clustered.

\section{Conclusions}\label{sec:conclusions}
We introduced a source-free domain adaptation (SFDA) method by uncovering the intrinsic target data structure. We proposed to achieve the adaptation by encouraging label consistency among local target features. We further considered density to reduce the negative impact of outliers. We differentiated between nearest neighbors, reciprocal neighbors and expanded neighborhood. Experimental results verified the importance of considering the local structure of the target features. Finally, our experimental results on both 2D image and 3D point cloud datasets testify the efficacy of our method.

\ifCLASSOPTIONcompsoc
  \section*{Acknowledgments}
\else
  \section*{Acknowledgment}
\fi

We acknowledge the support from Huawei Kirin Solution, and the project PID2022-143257NB-I00, financed by CIN/AEI/10.13039/501100011033 and FSE+, and Grant PID2021-128178OB-I00 funded by MCIN/AEI/ 10.13039/501100011033 and by ERDF A way of making Europe, Ramon y Cajal fellowship Grant RYC2019-027020-I funded by MCIN/AEI/ 10.13039/501100011033 and by ERDF A way of making Europe, and the CERCA Programme of Generalitat de Catalunya. Yaxing acknowledges the support from the project funded by Youth Foundation 62202243 (China).

\ifCLASSOPTIONcaptionsoff
  \newpage
\fi

{\small
\bibliography{longstrings,egbib}
\bibliographystyle{plain}
}

\begin{IEEEbiography}[{\includegraphics[width=1in,height=1.25in,clip,keepaspectratio]{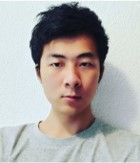}}]{Shiqi Yang}
joined Learning and Machine Perception (LAMP) team in 2019.10 as a Ph.D student advised by Dr. Joost van de Weijer in Computer Vision Center of Autonomous University of Barcelona, Spain. He received master degree in Huazhong University of Science and Technology, China, and once worked as a research associate in Kyoto University, Japan. His research interest focuses on
how to efficiently adapt the pretrained model to real
world environment under domain and category shift,
including source-free/continual/open-set/universal domain adaptation.
\end{IEEEbiography}

\begin{IEEEbiography}[{\includegraphics[width=1in,height=1.25in,clip,keepaspectratio]{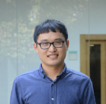}}]{Yaxing Wang}
is an associate professor of college of computer science at Nankai University. His research interests include GANs, image-to-image translation, domain adaptation and lifelong learning. Prior to joining NKU I was a postdoc at UAB, CVC, working with Joost van de Weije. I obtained my Ph.D. from Universitat Autònoma de Barcelona, under the supervision of Joost van de Weijer. I also experienced amazing internship at IIAI with Fahad shahbaz khan and Salman Khan.
\end{IEEEbiography}

\begin{IEEEbiography}[{\includegraphics[width=1in,height=1.25in,clip,keepaspectratio]{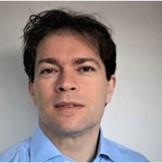}}]{Joost van de Weijer}
received the Ph.D. degree
from the University of Amsterdam, Amsterdam,
Netherlands, in 2005.
He was a Marie Curie Intra-European Fellow with INRIA Rhone-Alpes, France, and from
2008 to 2012, he was a Ramon y Cajal Fellow with
the Universitat Autònoma de Barcelona, Barcelona,
Spain, where he is currently a Senior Scientist
with the Computer Vision Center and leader of the
Learning and Machine Perception (LAMP) Team.
His main research directions are color in computer
vision, continual learning, active learning, and domain adaptation.
\end{IEEEbiography}

\begin{IEEEbiography}[{\includegraphics[width=1in,height=1.25in,clip,keepaspectratio]{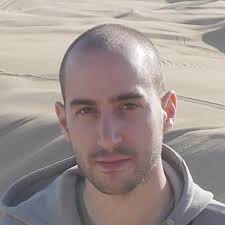}}]{Luis Herranz}
is a senior researcher with the Computer Vision Centre, and adjunct professor with the Universitat Autònoma de Barcelona (UAB). From 2012 to 2016 I worked with the Visual Information Processing and Learning (VIPL) of the Institute of Computing Technology (ICT) of the Chinese Academy of Sciences (CAS) in Beijing (China). Previously, I worked with Mitsubishi Electric R\&D Centre Europe in Guildford, United Kingdom, and with the Video Processing and Understanding Lab of the Escuela Politécnica Superior of the Universidad Autónoma de Madrid (UAM), where I received my Ph.D.
\end{IEEEbiography}

\begin{IEEEbiography}[{\includegraphics[width=1in,height=1.25in,clip,keepaspectratio]{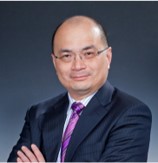}}]{Shangling Jui}
is the Chief AI Scientist for Huawei Kirin Chipset Solution. He is an expert in machine learning, deep learning, and artificial intelligence. Previously, he was the President of the SAP China Research Center and the SAP Korea Research Center. He was also the CTO of Pactera, leading innovation projects based on cloud and big data technologies. He received the Magnolia Award from the Municipal Government of Shanghai, in 2011.
\end{IEEEbiography}

\begin{IEEEbiography}[{\includegraphics[width=1in,height=1.25in,clip,keepaspectratio]{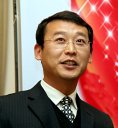}}]{Jian Yang}
(Member, IEEE) received the Ph.D.
degree in pattern recognition and intelligence systems from the Nanjing University of Science and
Technology (NUST), Nanjing, China, in 2002.
In 2003, he was a Post-Doctoral Researcher
with the University of Zaragoza, Zaragoza, Spain.
From 2004 to 2006, he was a Post-Doctoral Fellow
with the Biometrics Centre, The Hong Kong Polytechnic University, Hong Kong. From 2006 to 2007,
he was a Post-Doctoral Fellow with the Department
of Computer Science, New Jersey Institute of Technology, Newark, NJ, USA. From 2006 to 2007, he is a Chang-Jiang Professor
with the School of Computer Science and Engineering, NUST. He is the author
of more than 200 scientific papers in pattern recognition and computer vision.
His papers have been cited over 30 000 times in Google Scholar. His research
interests include pattern recognition, computer vision, and machine learning.
Dr. Yang is also a fellow of IAPR. He is/was an Associate Editor of Pattern
Recognition, Pattern Recognition Letters, IEEE TRANSACTIONS ON NEURAL
NETWORKS AND LEARNING SYSTEMS, and Neurocomputing.
\end{IEEEbiography}





\end{document}